\documentclass[10pt,twocolumn,letterpaper]{article}

\usepackage{cvpr}              

\usepackage{graphicx}
\usepackage{amsmath}
\usepackage{amssymb}
\usepackage{booktabs}
\usepackage{overpic}
\usepackage{amssymb}
\usepackage{times}
\usepackage{epsfig}
\usepackage{makecell}
\usepackage{caption}
\usepackage{float}
\usepackage{amsmath}
\usepackage{amssymb}
\usepackage{booktabs}
\usepackage{color}
\usepackage{amstext}
\usepackage{array}

\usepackage{setspace}
\usepackage{chngpage}
\usepackage{longtable}
\usepackage{footnote}
\usepackage{multirow}
\usepackage{makecell}
\usepackage{bigstrut}
\usepackage{tabularx}
\usepackage{bbding}
\usepackage{algorithm}
\usepackage{algorithmic}
\usepackage{bm}
\usepackage{bbding}
\usepackage{diagbox}
\usepackage[pagebackref,breaklinks,colorlinks]{hyperref}

\usepackage[capitalize]{cleveref}
\crefname{section}{Sec.}{Secs.}
\Crefname{section}{Section}{Sections}
\Crefname{table}{Table}{Tables}
\crefname{table}{Tab.}{Tabs.}

\def\cvprPaperID{4041} 
\def\confName{CVPR}
\def\confYear{2022}
\graphicspath{{figure/}}

\begin{document}

\title{Incorporating Semi-Supervised and Positive-Unlabeled Learning for Boosting Full Reference Image Quality Assessment}

	\author{Yue Cao$^1$, Zhaolin Wan$^2$, Dongwei Ren$^{1(}$\Envelope$^)$, Zifei Yan$^1$,  Wangmeng Zuo$^{1,3}$\\
	\\
		$^1$School of Computer Science and Technology, Harbin Institute of Technology, China\\
		$^2$College of Artificial Intelligence, Dalian Maritime University, China
		$^3$Peng Cheng Laboratory, China\\
		{\tt\small cscaoyue@gmail.com, zlwan@dlmu.edu.cn, rendongweihit@gmail.com, \{yanzifei, wmzuo\}@hit.edu.cn}
    }

\maketitle

\begin{abstract}
\vspace*{-2mm}
Full-reference (FR) image quality assessment (IQA) evaluates the visual quality of a distorted image by measuring its perceptual difference with pristine-quality reference, and has been widely used in low-level vision tasks.
%
%
Pairwise labeled data with mean opinion score (MOS) are required in training FR-IQA model, but is time-consuming and cumbersome to collect.
%
In contrast, unlabeled data can be easily collected from an image degradation or restoration process, making it encouraging to exploit unlabeled training data to boost FR-IQA performance.
%
%
%
%
%
Moreover, due to the distribution inconsistency between labeled and unlabeled data, outliers may occur in unlabeled data, further increasing the training difficulty.
%
%
In this paper, we suggest to incorporate semi-supervised and positive-unlabeled (PU) learning for exploiting unlabeled data while mitigating the adverse effect of outliers.
Particularly, by treating all labeled data as positive samples, PU learning is leveraged to identify negative samples (i.e., outliers) from unlabeled data.
%
%
Semi-supervised learning (SSL) is further deployed to exploit positive unlabeled data by dynamically generating pseudo-MOS.
%
%
%
We adopt a dual-branch network including reference and distortion branches.
Furthermore, spatial attention is introduced in the reference branch to concentrate more on the informative regions, and sliced Wasserstein distance is used for robust difference map computation to address the misalignment issues caused by images recovered by GAN models.
%
%
%
%
Extensive experiments show that our method performs favorably against  state-of-the-arts on the benchmark datasets PIPAL, KADID-10k, TID2013, LIVE and CSIQ.
The source code and model are available at {\url{{https://github.com/happycaoyue/JSPL}}}.
\end{abstract}
%
\vspace*{-2mm}
\section{Introduction}
\label{sec:intro}
\vspace*{-1mm}
The goal of image quality assessment is to provide computational models that can automatically predict the perceptual image quality consistent with human subjective perception.
Over the past few decades, significant progress has been made in developing full reference (FR) image quality assessment (IQA) metrics, including peak signal-to-noise ratio (PSNR) and structural similarity index (SSIM)~\cite{SSIM}, which have been widely used in various image processing fields.
Recently, CNN-based FR-IQA models have attracted more attention, which usually learn a mapping from distorted and pristine images to mean opinion score.
%

\begin{figure}[!t]
\scriptsize{
\begin{center}
\vspace{-0ex}
\begin{overpic}[width=0.42\textwidth,scale=0.5]{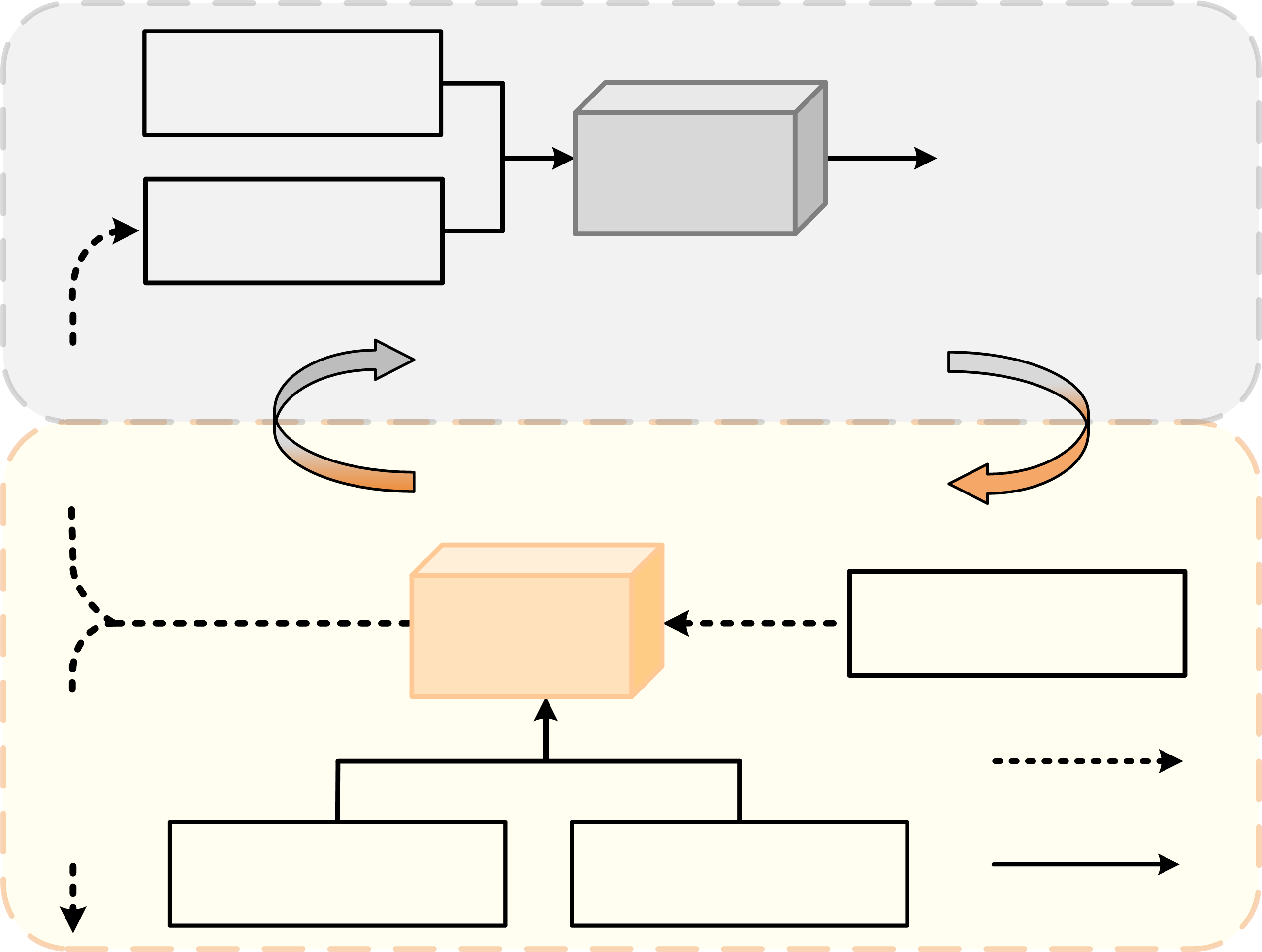} 
\put(18,69.5){\color{red}\scalebox{0.95}{$\mathrm{\textbf{Positive}}$}}
\put(13.5,65.5){\color{black}\scalebox{0.95}{$\mathrm{\textbf{Labeled \, Data}}$}}
\put(18,58){\color{red}\scalebox{0.95}{$ \mathrm{\textbf{Positive}}$}}
\put(12.1,54){\color{black}\scalebox{0.95}{$\mathrm{\textbf{Unlabeled \, Data}}$}}
\put(46.5,60.5){\color{black}\scalebox{0.95}{$\mathrm{\textbf{IQA  Model}}$}}
\put(74.9,61.8){\color{black}\scalebox{0.95}{$\mathrm{\textbf{Prediction  Score}}$}}

\put(35.5,45.5){\color{black}\scalebox{0.95}{$\mathrm{\textbf{Semi-supervised  Learning}}$}}
\put(34.3,35.9){\scalebox{0.95}{$\textcolor[RGB]{234,112,13}{\mathrm{\textbf{Positive-unlabeled  Learning}}}$}}

\put(36.9,26){\color{black}\scalebox{0.95}{$\mathrm{\textbf{Binary}}$}}
\put(35.3,22){\color{black}\scalebox{0.95}{$\mathrm{\textbf{Classifier}}$}}
\put(69.5,24.8){\color{black}\scalebox{0.95}{$\mathrm{\textbf{Unlabeled \, Data}}$}}
\put(21.4,6.8){\color{red}\scalebox{0.95}{$\mathrm{\textbf{Positive}}$}}
\put(17.5,3){\color{black}\scalebox{0.95}{$\mathrm{\textbf{Labeled \, Data}}$}}
\put(47.8,5){\color{black}\scalebox{0.95}{$\mathrm{\textbf{Unlabeled \, Data}}$}}

\put(4.4,36.5){\color{red}\scalebox{0.95}{$\mathrm{\rotatebox{90}{\textbf{Positive}}}$}}
\put(4.4,8.2){\scalebox{0.95}{$\textcolor[RGB]{112,48,160}{\mathrm{\rotatebox{90}{\textbf{Negative}}}}$}}

\put(79.8,11.5){\color{black}\scalebox{0.95}{$\mathrm{\textbf{Inference}}$}}
\put(79.8,3){\color{black}\scalebox{0.95}{$\mathrm{\textbf{Training}}$}}

\end{overpic}
\end{center}}
\vspace{-4.5ex}
\caption{\small
{Illustration of joint semi-supervised and PU learning (JSPL) method, which mitigates the adverse effect of outliers in unlabeled data for boosting the performance of IQA model.
}
}
\vspace*{-4mm}
\label{fig:intro}
\end{figure}
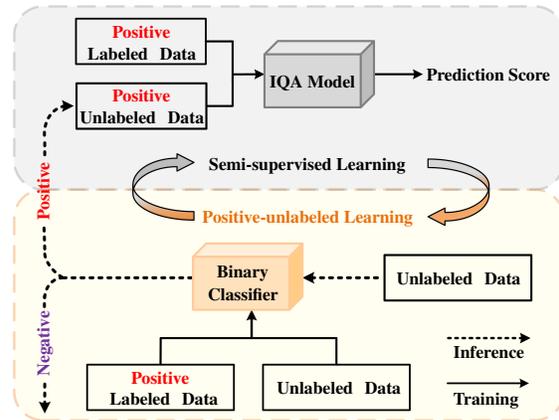

Most existing CNN-based FR-IQA models are trained using pairwise labeled data with mean opinion score (MOS), thus requiring extensive human judgements.
%
%
To reduce the cost of collecting a large amount of labeled data, a potential alternative is semi-supervised learning for exploiting unlabeled samples which are almost free.
%
%
Recently, considerable attention has been given to semi-supervised IQA algorithms~\cite{tang2014blind, wang2021semi, xu2020semi, liu2020real, liu2020hierarchical} which show promising performance using both labeled and unlabeled data.
%
%
%
%
%
However, unlabeled data can be collected in various unconstrained ways and may have a much different distribution from labeled data.
%
%
Consequently, outliers usually are inevitable and are harmful to semi-supervised learning~\cite{guo2020safe}.
%
%

In this paper, we incorporate semi-supervised and positive-unlabeled (PU) learning for exploiting unlabeled data while mitigating the adverse effect of outliers.
%
%
PU learning aims at learning a binary classifier from a labeled set of positive samples as well as an unlabeled set of both positive and negative samples, and has been widely applied in image classification~\cite{Self-PU} and anomaly detection~\cite{zhang2017positive}.
%
%
%
%
%
As for our task, the labeled images with MOS annotations can be naturally treated as positive samples.
As shown in Fig.~\ref{fig:intro}, PU learning is then exploited to find and exclude outliers, \ie, negative samples, from the unlabeled set of images without MOS annotations.
Then, semi-supervised learning (SSL) is deployed to leverage both labeled set and positive unlabeled images for training deep FR-IQA models.
Moreover, the prediction by PU learning can also serve as the role of confidence estimation to gradually select valuable positive unlabeled images for SSL.
Thus, our joint semi-supervised and PU learning (JSPL) method provides an effective and convenient way to incorporate both labeled and unlabeled sets for boosting FR-IQA performance.

Besides, we also present a new FR-IQA network for emphasizing informative regions and suppressing the effect of misalignment between distorted and pristine images.
Like most existing methods, our FR-IQA network involves a Siamese (\ie, dual-branch) feature extraction structure respectively for distorted and pristine images.
The pristine and distortion features are then fed into the distance calculation module to generate the difference map, which is propagated to the score prediction network to obtain the prediction score.
%
%
%
However, for GAN-based image restoration, the distorted image is usually spatially misaligned with the pristine image, making pixel-wise Euclidean distance unsuitable for characterizing the perceptual quality of distorted image~\cite{PIPAL,SWDN}.
To mitigate this, Gu~\cite{SWDN} introduced a pixel-wise warping operation, \ie, space warping difference (SWD).
In this work, we extend sliced Wasserstein distance to its local version (LocalSW) for making the difference map robust to small misalignment while maintaining its locality.
%
%
%
%
%
%
%
Moreover, human visual system (HVS) usually pays more visual attention to the image regions containing more informative content~\cite{simoncelli2001natural, najemnik2005optimal,IW-SSIM, CSIQ}, and significant performance improvements have been achieved by considering the correlation with human visual fixation or visual region-of-interest detection~\cite{larson2008unveiling, larson2008can, engelke2008regional}.
%
%
Taking the properties of HVS into account, we leverage spatial attention modules on pristine feature for emphasizing more on informative regions, which are then used for reweighting distance map to generate the calibrated difference maps.

Extensive experiments are conducted to evaluate our JSPL method for FR-IQA.
Based on the labeled training set, we collect unlabeled data by using several representative image degradation or restoration models.
On the Perceptual Image Processing ALgorithms  (PIPAL) dataset~\cite{PIPAL}, the results show that both JSPL, LocalSW, and spatial attention contribute to performance gain of our method, which performs favorably against state-of-the-arts for assessing perceptual quality of GAN-based image restoration results.
%
%
%
We further conduct experiments on four traditional IQA datasets, \ie, LIVE~\cite{LIVE}, CSIQ~\cite{CSIQ}, TID2013~\cite{TID2013} and KADID-10k~\cite{KADID}, further showing the superiority of our JSPL method against state-of-the-arts.

To sum up, the main contribution of this work includes:
\begin {itemize}
   \vspace*{-2.5mm}
   \item A joint semi-supervised and PU learning (JSPL) method is presented to exploit images with and without MOS annotations for improving FR-IQA performance.
   In comparison to SSL, PU learning plays a crucial role in our JSPL by excluding outliers and gradually selecting positive unlabeled data for SSL.
   %
   %
   \vspace*{-2.5mm}
   \item In FR-IQA network, spatial attention and local sliced Wasserstein distance are further deployed in computing difference map for emphasizing informative regions and suppressing the effect of misalignment between distorted and pristine image.
   %
   %
   %
   %
   \vspace*{-2.5mm}
   \item Extensive experiments on five benchmark IQA datasets show that our JSPL model performs favorably against the state-of-the-art FR-IQA models.
 \end {itemize}

 \section{Related Work}
\label{sec:relate}
\vspace*{-1mm}
In this section, we present a brief review on learning-based FR-IQA, semi-supervised IQA, as well as IQA for GAN-based image restoration.
%
%

\vspace*{-1mm}
\subsection{Learning-based FR-IQA Models}
\vspace*{-1mm}
Depending on the accessibility to the pristine-quality reference, IQA methods can be classified into full reference (FR), reduced reference (RR) and no reference (NR) models.
FR-IQA methods compare the distorted image against its pristine-quality reference, which can be further divided into two categories: traditional evaluation metrics and CNN-based models.
The traditional metrics are based on a set of prior knowledge related to the properties of HVS. However, it is difficult to simulate the HVS with limited hand-crafted features because visual perception is a complicated process.
In contrast, learning-based FR-IQA models use a variety of deep networks to extract features from training data without expert knowledge.

For deep FR-IQA, Gao \etal~\cite{gao2017deepsim} first computed the local similarities of the feature maps from VGGNet layers between the reference and distorted images. Then, the local similarities are pooled together to
get the final quality score.
DeepQA~\cite{DeepQA} applied CNN to regress the sensitivity map to subjective score, which was generated from distorted images and error maps.
Bosse \etal~\cite{WaDIQaM} presented a CNN-based FR-IQA method, where the perceptual image quality is obtained by weighted pooling on patch-wise scores.
Learned Perceptual Image Patch Similarity (LPIPS)~\cite{LPIPS} computed the Euclidean distance between reference and distorted deep feature representations, and can be flexibly embedded in various pre-trained CNNs, such as VGG~\cite{VGG} and AlexNet~\cite{AlexNet}.
Benefiting from SSIM-like structure and texture similarity measures, Ding \etal ~\cite{DISTS} presented a Deep Image Structure and Texture Similarity metric (DISTS) based on an injective mapping function.
Hammou \etal~\cite{EGB} proposed an ensemble of gradient boosting (EGB) metric based on selected feature similarity and ensemble learning.
Ayyoubzadeh \etal~\cite{ASNA} used Siamese-Difference neural network equipped with the spatial and channel-wise attention to predict the quality score.
All the above metrics require a large number of labeled images to train the model. However, manual labeling is expensive and time-consuming, making it appealing to better leverage unlabeled images for boosting IQA performance.
%
%

\vspace*{-1mm}
\subsection{Semi-Supervised IQA}
\vspace*{-1mm}
In recent years, semi-supervised IQA algorithms have attracted considerable attention, as they use less expensive and easily accessible unlabeled data, and are beneficial to performance improvement~\cite{cozman2003semi}.
%
%
Albeit semi-supervised learning (SSL) has been extensively studied and applied in vision and learning tasks, the research on semi-supervised IQA is still in its infancy.
%
%
Tang \etal~\cite{tang2014blind} employed deep belief network for IQA task, and the method was pre-trained with unlabeled data and then finetuned with labeled data.
Wang \etal~\cite{wang2021semi} utilized the semi-supervised ensemble learning for NR-IQA by combining labeled and unlabeled data, where unlabeled data is incorporated for maximizing ensemble diversity.
Lu \etal~\cite{lu2015blind} introduced semi-supervised local linear embedding (SS-LLE) to map the image features to the quality scores.
Zhao \etal~\cite{zhao2019face} proposed a SSL-based face IQA method, which exploits the unlabeled data in the target domain to finetune the network by predicting and updating labels.
In the field of medical imaging, the amount of labeled data is limited, and the annotated labels are highly private. And SSL~\cite{xu2020semi, liu2020real, liu2020hierarchical} provided an encouraging solution to address this problem by incorporating the unlabeled data with the labeled data to achieve better medical IQA performance.
 Nonetheless, the above studies assume that the labeled and unlabeled data are from the same distribution.
 %
 %
 %
 %
 However, the inevitable distribution inconsistency and outliers are harmful to SSL~\cite{guo2020safe}, but remain less investigated in semi-supervised IQA. %

\vspace*{-1mm}
\subsection{IQA for GAN-based Image Restoration}
\vspace*{-1mm}
Generative adversarial networks (GAN) have been widely adopted in image restoration for improving visual performance of restoration results.
However, these images usually suffer from texture-like artifacts aka GAN-based distortions that are seemingly fine-scale yet fake details.
Moreover, GAN is prone to producing restoration results with spatial distortion and misalignment, which also poses new challenges to existing IQA methods.
%
%
Recently, some intriguing studies have been proposed to improve the performance on IQA for GAN-based image restoration.
%
%
SWDN~\cite{SWDN} proposed a pixel-wise warping operation named space warping difference (SWD) to alleviate the spatial misalignment, by comparing the features within a small range around the corresponding position.
%
%
%
%
Shi \etal~\cite{RADN} deployed the reference-oriented deformable convolution and a patch-level attention module in both reference and distortion branches for improving the IQA performance on GAN-based distortion.
%
%
%
For modeling the GAN-generated texture-like noises, IQMA~\cite{IQMA} adopted a multi-scale architecture to measure distortions, and evaluated images at a fine-grained texture level.
IQT~\cite{IQT} combined CNN and transformer for IQA task, and achieved state-of-the-art performance.
%
%
Although progress has been made in evaluating GAN-based distortion, existing methods are based on labeled data via supervised learning.
In comparison, this work suggests a joint semi-supervised and PU learning method as well a new IQA network for leveraging unlabeled data and alleviating the spatial misalignment issue.


%
\section{Proposed Method}
\label{sec:method}

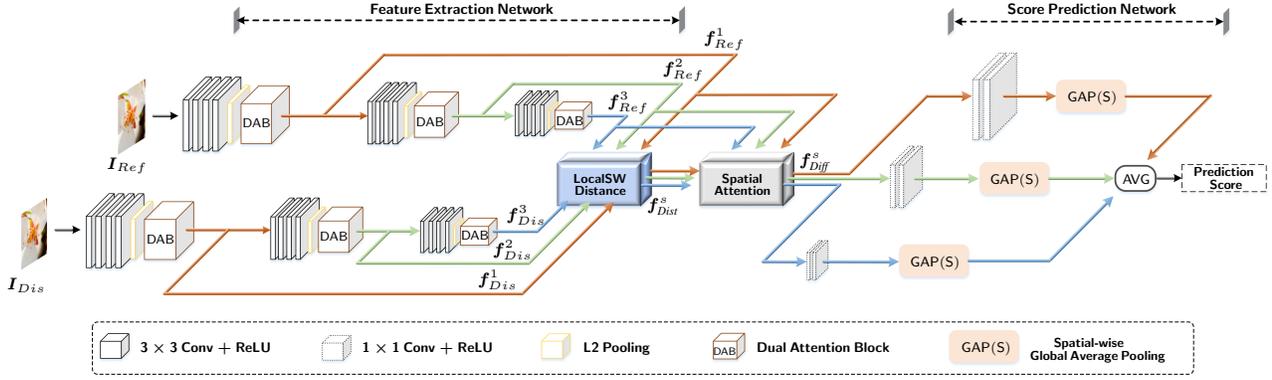
\begin{figure*}[!t]
\scriptsize{
\begin{center}
\vspace{-0ex}
\begin{overpic}[width=0.95\textwidth,scale=0.5]{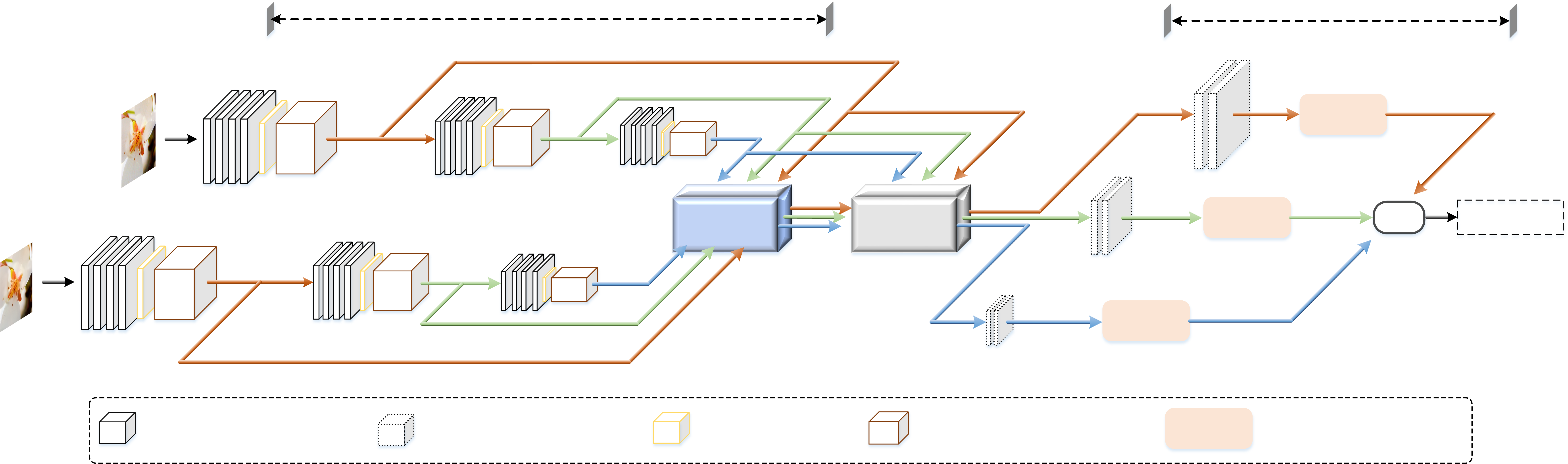} 
\put(6.5,16.5){\color{black}\scalebox{0.9}{ $\bm{I}_{Ref}$}}
\put(54.3,26.6){\color{black}\scalebox{0.9}{ $\bm{f}^{1}_{Ref}$}}
\put(51,24.1){\color{black}\scalebox{0.9}{ $\bm{f}^{2}_{Ref}$}}
\put(46.6,21.5){\color{black}\scalebox{0.9}{ $\bm{f}^{3}_{Ref}$}}

\put(-1.5,6.8){\color{black}\scalebox{0.9}{ $\bm{I}_{Dis}$}}
\put(36.2,7.2){\color{black}\scalebox{0.9}{ $\bm{f}^{1}_{Dis}$}}
\put(37.4,9.6){\color{black}\scalebox{0.9}{ $\bm{f}^{2}_{Dis}$}}
\put(38.5,12.4){\color{black}\scalebox{0.9}{ $\bm{f}^{3}_{Dis}$}}

\put(50.1,13.2){\color{black}\scalebox{0.9}{$\bm{f}^{s}_{\emph{Dist}}$}}
\put(62.3,16.8){\color{black}\scalebox{0.9}{$\bm{f}^{s}_{\emph{Diff}}$}}

\put(17.7,19.8){\color{black}\scalebox{0.7}{$\mathsf{DAB}$}}
\put(31.6,19.9){\color{black}\scalebox{0.7}{$\mathsf{DAB}$}}
\put(42.8,20){\color{black}\scalebox{0.7}{$\mathsf{DAB}$}}

\put(10,10.5){\color{black}\scalebox{0.7}{$\mathsf{DAB}$}}
\put(23.9,10.7){\color{black}\scalebox{0.7}{$\mathsf{DAB}$}}
\put(35.2,10.8){\color{black}\scalebox{0.7}{$\mathsf{DAB}$}}

\put(44.2,15.4){\color{black}\scalebox{0.7}{$\bm {\mathsf{LocalSW}}$}}
\put(44.3,14.4){\color{black}\scalebox{0.7}{$\bm {\mathsf{Distance}}$}}
\put(56.1,15.4){\color{black}\scalebox{0.7}{$\bm {\mathsf{Spatial}}$}}
\put(55.5,14.4){\color{black}\scalebox{0.7}{$\bm {\mathsf{Attention}}$}}

\put(83.8,21.9){\color{black}\scalebox{0.75}{$\mathsf{GAP(S)}$}}
\put(77.6,15.3){\color{black}\scalebox{0.75}{$\mathsf{GAP(S)}$}}
\put(71.2,8.7){\color{black}\scalebox{0.75}{$\mathsf{GAP(S)}$}}
\put(88.2,15.3){\color{black}\scalebox{0.75}{$\mathsf{AVG}$}}
\put(93.9,15.9){\color{black}\scalebox{0.7}{$\bm {\mathsf{Prediction}}$}}
\put(95.2,14.9){\color{black}\scalebox{0.7}{$\bm {\mathsf{Score}}$}}

\put(9.5,1.8){\color{black}\scalebox{0.75}{$\bm {\mathsf{3 \times 3 \ Conv + ReLU}}$}}
\put(27.3,1.8){\color{black}\scalebox{0.75}{$\bm {\mathsf{1 \times 1 \ Conv + ReLU}}$}}
\put(45,1.8){\color{black}\scalebox{0.75}{$\bm {\mathsf{L2 \ Pooling}}$}}
\put(55.4,1.6){\color{black}\scalebox{0.58}{$\mathsf{DAB}$}}
\put(58.9,1.8){\color{black}\scalebox{0.75}{$\bm {\mathsf{Dual \ Attention \ Block}}$}}
\put(75.3,1.8){\color{black}\scalebox{0.75}{$\mathsf{GAP(S)}$}}
\put(82.7,2.2){\color{black}\scalebox{0.7}{$\bm {\mathsf{Spatial\textbf{-} wise}}$}}
\put(80.8,1.2){\color{black}\scalebox{0.7}{$\bm {\mathsf{Global \, Average \, Pooling}}$}}

\put(27.9,28.9){\color{black}\scalebox{0.8}{$\bm {\mathsf{Feature  \, Extraction \,  Network}}$}}
\put(79,28.9){\color{black}\scalebox{0.8}{$\bm {\mathsf{Score \,  Prediction \,  Network}}$}}

\end{overpic}

\caption{\small
{Illustration of our FR-IQA network.
It adopts a dual-branch structure for feature extraction, \ie, one for reference and another for distortion.
The feature extraction network performs feature extraction on reference and distortion images at three scales.
The distance calculation module generates the difference map between the above two features.
The spatial attention module gives greater weight on more informative regions to obtain the calibrated difference map, which is then fed into score prediction network to predict the final score.}}
\label{fig:arch}
\vspace{-7mm}
\end{center}}
\end{figure*}
\subsection{Problem Setting}

Denote by $\bm{x} = (\bm{I}_{Ref},\bm{I}_{Dis})$ a two-tuple of pristine-quality reference image $\bm{I}_{Ref}$ and distorted image $\bm{I}_{Dis}$, and $y$ the ground-truth MOS.
Learning-based FR-IQA aims to find a mapping $f(\bm{x})$ parameterized by $\Theta^{f}$ to predict the quality score $\hat{y}$ for approximating $y$.
Most existing FR-IQA methods are based on supervised learning where the collection of massive MOS annotations is very time-consuming and cumbersome.
In this work, we consider a more encouraging and practically feasible SSL setting, \ie, training FR-IQA model using labeled data as well as unlabeled data with outliers.
While SSL has been suggested to exploit unlabeled data for boosting IQA performance, we note that outliers usually are inevitable when unlabeled data are collected with diverse and unconstrained ways.
For example, reference image quality of some unlabeled two-tuples may not meet the requirement. %
And the unlabeled data may also contain distortion types unseen in labeled data and non-necessary for IQA training.

Let $\mathbb{P}=\left\{\bm{x}_{i}, y_{i}\right\}_{i=1}^{N_{p}}$ denote the positive labeled data and $\mathbb{U}=\left\{\bm{x}_{j}\right\}_{j=1}^{N_{u}}$ denote unlabeled data.
We present a joint semi-supervised and PU learning (JSPL) method for leveraging the unlabeled data with potential outliers.
%
%
%
%
%
Besides the IQA model $f(\bm{x})$, our JSPL also learns a binary classifier $h(\bm{x}_j)$ parameterized by $\Theta^{h}$ for determining an unlabeled two-tuple is a negative (\ie, outlier) or a positive sample.

%

\subsection{JSPL Model}
A joint semi-supervised and PU learning (JSPL) model is presented to learn IQA model $f(\bm{x})$ and binary classifier $h(\bm{x})$ from the labeled data $\mathbb{P}$ and the unlabeled data $\mathbb{U}$.
Particularly, PU learning is utilized to learn $h(\bm{x})$ for identifying positive unlabeled samples.
And SSL is used to learn $f(\bm{x})$ from both labeled and positive unlabeled samples.
In the following, we first describe the loss terms for PU learning and SSL, and then introduce our overall JSPL model.

\textbf{PU Learning.} In order to learn $h(\bm{x})$, we treat all samples in $\mathbb{P}$ as positive samples, and all samples in $\mathbb{U}$ as unlabeled samples.
For a positive sample $\bm{x}_i$, we simply adopt the cross-entropy (CE) loss,
\begin{equation}
\label{eq:pce}
CE(h(\bm{x}_i)) = -\log h(\bm{x}_i).
\end{equation}
Each unlabeled sample $\bm{x}_j$ should be either positive or negative sample, and we thus require the output $h(\bm{x}_j)$ to approach either 1 or 0.
To this end, we introduce the entropy loss defined as,
\begin{equation}
\small
\label{eq:entropy}
\setlength{\abovedisplayskip}{4.5pt}
\setlength{\belowdisplayskip}{4.5pt}
\mathcal{H}(h(\bm{x}_j)) \!=\! - h(\bm{x}_j) \log h(\bm{x}_j) \!-\! (1 \!-\! h(\bm{x}_j)) \log (1 \!-\! h(\bm{x}_j)).
\end{equation}
We note that the entropy loss has been widely used in SSL~\cite{entropy_min}.
When only using CE loss and entropy loss, $h(\bm{x})$ may simply produce 1 for any sample $\bm{x}$.
To tackle this issue, for a given mini-batch $\mathcal{B}_u$ of unlabeled samples, we introduce a negative-enforcing (NE) loss for constraining that there is at least one negative sample in each mini-batch,
\begin{equation}
\label{eq:neloss}
\setlength{\abovedisplayskip}{3.5pt}
\setlength{\belowdisplayskip}{3.5pt}
NE(\mathcal{B}_u) = - \log \left( 1 - \min\nolimits_{\bm{x}_j \in \mathcal{B}_u} h(\bm{x}_j)\right).
\end{equation}
Combining the above loss terms, we define the PU learning loss as,
\begin{equation}
\small
\setlength{\abovedisplayskip}{3.5pt}
\setlength{\belowdisplayskip}{3.5pt}
\label{eq:PU}
\begin{split}
\mathcal{L}_{\emph{PU}}\!\!=\!\!\sum\nolimits_{i}\! CE(h(\bm{x}_i))\!+\!\sum\nolimits_{j}\! \mathcal{H}\left(h(\bm{x}_{j})\right)\!+\! \sum\nolimits_{\mathcal{B}_u}\! NE(\mathcal{B}_u).
\end{split}
\end{equation}

\textbf{SSL.} FR-IQA is a regression problem.
For labeled sample $\bm{x}_i$ with ground-truth MOS $y_i$, we adopt the mean squared error (MSE) loss defined as,
\begin{equation}
\label{eq:mse}
\ell(f(\bm{x}_i), y_i) = \| f(\bm{x}_i) - y_i \|^2.
\end{equation}
As for unlabeled data, only the positive unlabeled samples (\ie, $h(\bm{x}_j) \geq \tau$) are considered in SSL.
Here, $\tau$ (\eg, $=0.5$) is a threshold for selecting positive unlabeled samples.
For positive unlabeled samples, we also adopt the MSE loss,
\begin{equation}
\label{eq:pseudomse}
\setlength{\abovedisplayskip}{5.0pt}
\setlength{\belowdisplayskip}{5.0pt}
\ell(f(\bm{x}_j), y^{*}_j) = \| f(\bm{x}_j) - y^{*}_j \|^2,
\end{equation}
where $y^{*}_j$ denotes the pseudo MOS for $\bm{x}_j$.
In SSL, sharpening is usually used for classification tasks to generate the pseudo label for unlabeled samples~\cite{MixMatch,FixMatch}, but is not suitable for regression tasks.
Motivated by~\cite{ELR,TemporalEnsembling}, we use the moving average strategy to obtain $y^{*}_j$ during training,
\begin{equation}
\setlength{\abovedisplayskip}{5.0pt}
\setlength{\belowdisplayskip}{5.0pt}
\label{eq:pseudomos}
\begin{split}
y_{j}^{*}(t)=\alpha \cdot y_{j}^{*}(t-1)+(1-\alpha) \cdot f^{t}\left(\bm{x}_{j}\right),
\end{split}
\end{equation}
where $\alpha$ ($=0.95$) is the momentum.
$y_{j}^{*}(t)$ denotes the pseudo MOS after $t$ iterations of training, and $f^{t}(\bm{x}_{j})$ denotes the network output after $t$ iterations of training.
Therefore, we define the SSL loss as,
\begin{equation}
\small
\setlength{\abovedisplayskip}{7.0pt}
\setlength{\belowdisplayskip}{7.0pt}
\label{eq:IQA}
\begin{split}
\mathcal{L}_{\emph{SSL}}\!=\!\sum\nolimits_{i}\! \ell\left(f(\bm{x}_{i}), y_{i}\right)\!+\!\sum\nolimits_{j} \mathbb{I}_{h(\bm{x}_{j}) \geq \tau}  \ell\left(f(\bm{x}_{j}), y_{j}^{*}\right).
\end{split}
\end{equation}
$\mathbb{I}_{h(\bm{x}_{j}) \geq \tau} $ is an indicator function, where it is 1 if ${h(\bm{x}_{j}) \geq \tau} $ and 0 otherwise.

\textbf{JSPL Model.} Taking the losses for both SSL and PU learning into account, the learning objective for JSPL can be written as,
\begin{equation}
\setlength{\abovedisplayskip}{4.0pt}
\setlength{\belowdisplayskip}{4.0pt}
\label{eq:objective}
\begin{split}
\min _{\Theta^{f}, \Theta^{h}} \mathcal{L} = \mathcal{L}_{SSL} + \mathcal{L}_{PU}.
\end{split}
\end{equation}
We note that our JSPL is a joint learning model, where both the FR-IQA network $f(\bm{x})$ and binary classifier $h(\bm{x})$ can be learned by minimizing the above objective function.
Particularly, for a given mini-batch of unlabeled samples, we first update the binary classifier by minimizing $\mathcal{L}_{PU}$.
Then, pseudo MOS is updated for each unlabeled sample, and positive unlabeled samples are selected.
Furthermore, the positive unlabeled samples are incorporated with the mini-batch of labeled samples to update the FR-IQA network by minimizing $\mathcal{L}_{SSL}$.

\subsection{FR-IQA Network Structure}

As shown in Fig.~\ref{fig:arch}, our proposed FR-IQA consists of a feature extraction network and a score prediction network.
The feature extraction network adopts a Siamese (\ie, dual-branch) structure, which respectively takes the reference image and the distorted image as the input.
It is based on VGG16~\cite{VGG} consisting of three different scales, \ie, $s=1$, $2$ and $3$.
And we further modify the VGG16 network from two aspects.
First, all max pooling layers in VGG are replaced with $L_2$ pooling~\cite{henaff2015geodesics} to avoid aliasing when downsampling by a factor of two.
Second, to increase the fitting ability, dual attention blocks (DAB) used in~\cite{CycleISP} are integrated into different scales of backbone network.
The reference image $\bm{I}_{Ref}$ and distorted image $\bm{I}_{Dis}$ are fed into the feature extraction network to obtain the reference feature $\bm{f}^{s}_{Ref}$ and  distortion feature $\bm{f}^{s}_{Dis}$ ($s = 1, 2, 3$), respectively.
Then, local sliced Wasserstein (LocalSW) distance is presented to produce distance map $\bm{f}^{s}_\emph{{Dist}}$, and a spatial attention module is deployed for reweighting distance map to generate calibrated difference map $\bm{f}^{s}_{\emph{Diff}}$ for each scale $s$.
As shown in Fig.~\ref{fig:arch}, the score prediction network has three branches, where each branch involves two $1\times1$ convolutional layers and a spatial-wise global averaging pooling layer.
$\bm{f}^{s}_\emph{{Diff}}$ is fed to the $s$-th branch to generate the score at scale $s$, and the scores at all scales are averaged to produce the final score.

In the following, we elaborate more on the LocalSW distance and difference map calibration.

%
%
%
%
%
%
%
%
%

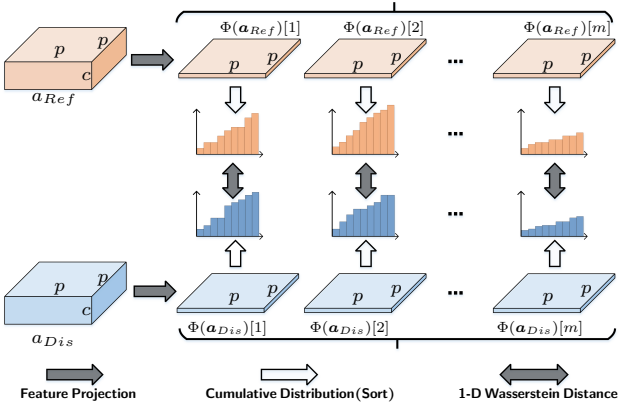
\begin{figure}[!t]
\scriptsize{
\begin{center}
\vspace{-0ex}
\begin{overpic}[width=0.47\textwidth,scale=0.5]{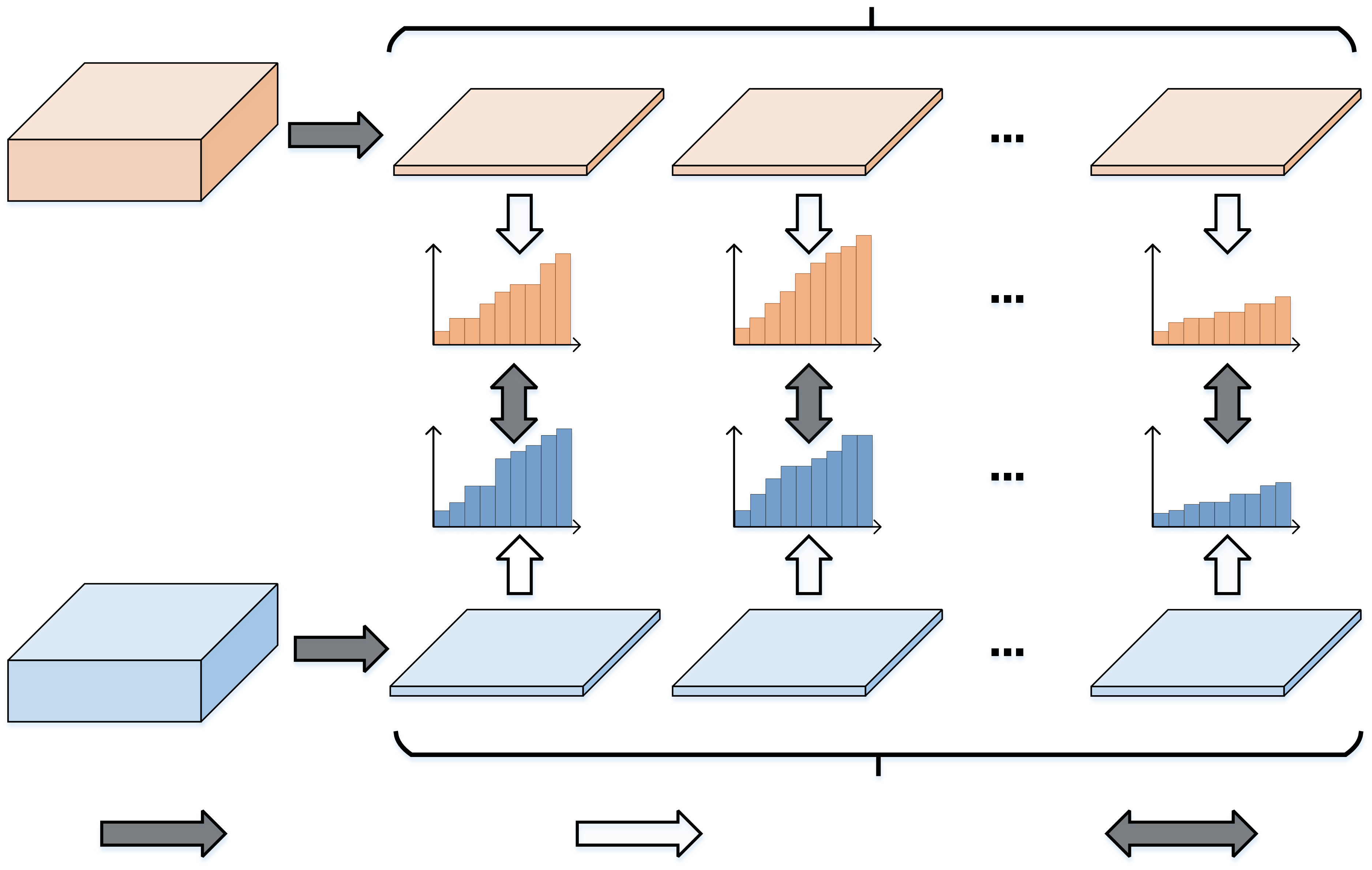} 
\put(4.8,47){\color{black}{${a}_{Ref}$}}
\put(35.3,58.1){\color{black}{\scalebox{0.80}{$\Phi({\bm{a}}_{Ref}){[1]}$}}}
\put(55.3,58.1){\color{black}{\scalebox{0.80}{$\Phi({\bm{a}}_{Ref}){[2]}$}}}
\put(84.5,58.1){\color{black}{\scalebox{0.80}{$\Phi({\bm{a}}_{Ref}){[m]}$}}}

\put(4.8,7.7){\color{black}{${a}_{Dis}$}}
\put(30.1,9.5){\color{black}{\scalebox{0.80}{$\Phi({\bm{a}}_{Dis}){[1]}$}}}
\put(49.9,9.5){\color{black}{\scalebox{0.80}{$\Phi({\bm{a}}_{Dis}){[2]}$}}}
\put(80.2,9.5){\color{black}{\scalebox{0.80}{$\Phi({\bm{a}}_{Dis}){[m]}$}}}

\put(8.5,54.5){\color{black}{$\tiny{p}$}}
\put(15.5,56){\color{black}{$\tiny{p}$}}
\put(12.9,50.4){\color{black}{$\tiny{c}$}}

\put(36.8,52.1){\color{black}{$\tiny{p}$}}
\put(43,53.4){\color{black}{$\tiny{p}$}}

\put(56.8,52.1){\color{black}{$\tiny{p}$}}
\put(63,53.4){\color{black}{$\tiny{p}$}}

\put(87.6,52.1){\color{black}{$\tiny{p}$}}
\put(93.8,53.4){\color{black}{$\tiny{p}$}}

\put(8.5,16.1){\color{black}{$\tiny{p}$}}
\put(15.5,17.6){\color{black}{$\tiny{p}$}}
\put(12.9,12.2){\color{black}{$\tiny{c}$}}

\put(36.8,14.5){\color{black}{$\tiny{p}$}}
\put(43,15.8){\color{black}{$\tiny{p}$}}

\put(56.8,14.5){\color{black}{$\tiny{p}$}}
\put(63,15.8){\color{black}{$\tiny{p}$}}

\put(87.6,14.5){\color{black}{$\tiny{p}$}}
\put(93.8,15.8){\color{black}{$\tiny{p}$}}

\put(3,-1.2){\color{black}\scalebox{0.75}{$\bm {\mathsf{Feature \, Projection}}$}}
\put(33,-1.2){\color{black}\scalebox{0.75}{$\bm {\mathsf{Cumulative \, Distribution(Sort)}}$}}
\put(74,-1.2){\color{black}\scalebox{0.75}{$\bm {\mathsf{1\textbf{-}D \, Wasserstein \, Distance}}$}}

\end{overpic}
\end{center}}
\vspace{-4.5ex}
\caption{\small
{The proposed local sliced Wasserstein distance (LocalSW) calculation module which measures the 1-D Wasserstein distance between cumulative distribution of the projected reference and distortion feature maps.}
}
\vspace*{-4mm}
\label{fig:SW}
\end{figure}

%
\textbf{LocalSW Distance.}~Given the reference feature $\bm{f}^{s}_{Ref}$ and  distortion feature $\bm{f}^{s}_{Dis}$, one direct solution is the element-wise difference, \ie,  $| \bm{f}^{s}_{Ref} - \bm{f}^{s}_{Dis} |$.
Here $|\cdot|$ denotes element-wise absolute value.
However, GAN-based restoration is prone to producing results being spatially distorted and misaligned with the reference image, while the element-wise difference is not robust to spatial misalignment.
Instead, we suggest local sliced Wasserstein (LocalSW) distance which measures the difference by comparing the distributions of feature maps.
%
%
%
%
%
Previously sliced Wasserstein loss~\cite{projected, heitz2021sliced} has been proposed to calculate the global sliced Wasserstein distance.
Considering that the misalignment between $\bm{f}^{s}_{Ref}$ and  $\bm{f}^{s}_{Dis}$ is usually local and within a small range, we adopt LocalSW distance by dividing $\bm{f}^{s}_{Ref}$ and $\bm{f}^{s}_{Dis}$ ($\in \mathbb{R}^{H \times W \times C}$) into $J$ non-overlapped patches with resolution $p \times p$, \ie, $J=(H/p)\times(W/p)$.
%
%
%
%
%
Fig.~\ref{fig:SW} illustrates the computation of LocalSW distance by using a patch pair $\bm{a}_{Ref}$ and $\bm{a}_{Dis}$ $(\in \mathbb{R}^{p \times p \times C})$ as an example.
In particular, we first use the projection operator $\Phi$ on $\bm{a}_{Ref}$ and $\bm{a}_{Dis}$ to obtain the projected features $\Phi(\bm{a}_{Ref})$ and $\Phi(\bm{a}_{Dis})$ $(\in \mathbb{R}^{p \times p \times m})$, where $m = C / 2$.
Then, we implement the cumulative distributions through sorting operation $\mathbf{Sort}(\cdot)$ on each channel (\ie, slice) $v$ of $\Phi(\bm{a}_{Ref})$ and $\Phi(\bm{a}_{Dis})$.
And the LocalSW distance for slice $v$ of this patch pair can be obtained by,
\begin{equation}
\small
\setlength{\abovedisplayskip}{5pt}
\setlength{\belowdisplayskip}{5pt}
\label{eq:LSW}
\begin{aligned}
\mathcal{SW}[v]= \!\!\left\|\mathbf{Sort}(\Phi(\bm{a}_{Ref})[v])\!-\!\mathbf{Sort}(\Phi(\bm{a}_{Dis})[v])\right\|.
\end{aligned}
\end{equation}
%
%
Furthermore, we compute the LocalSW distance for all slices and all patches to form the LocalSW distance map $\bm{f}^{s}_{Dist} \in \mathbb{R}^{ \frac{H}{p} \times \frac{W}{p} \times m}$.


\begin{figure}[!t]
\scriptsize{
\begin{center}
\vspace{-0ex}
\begin{overpic}[width=0.47\textwidth,scale=0.5]{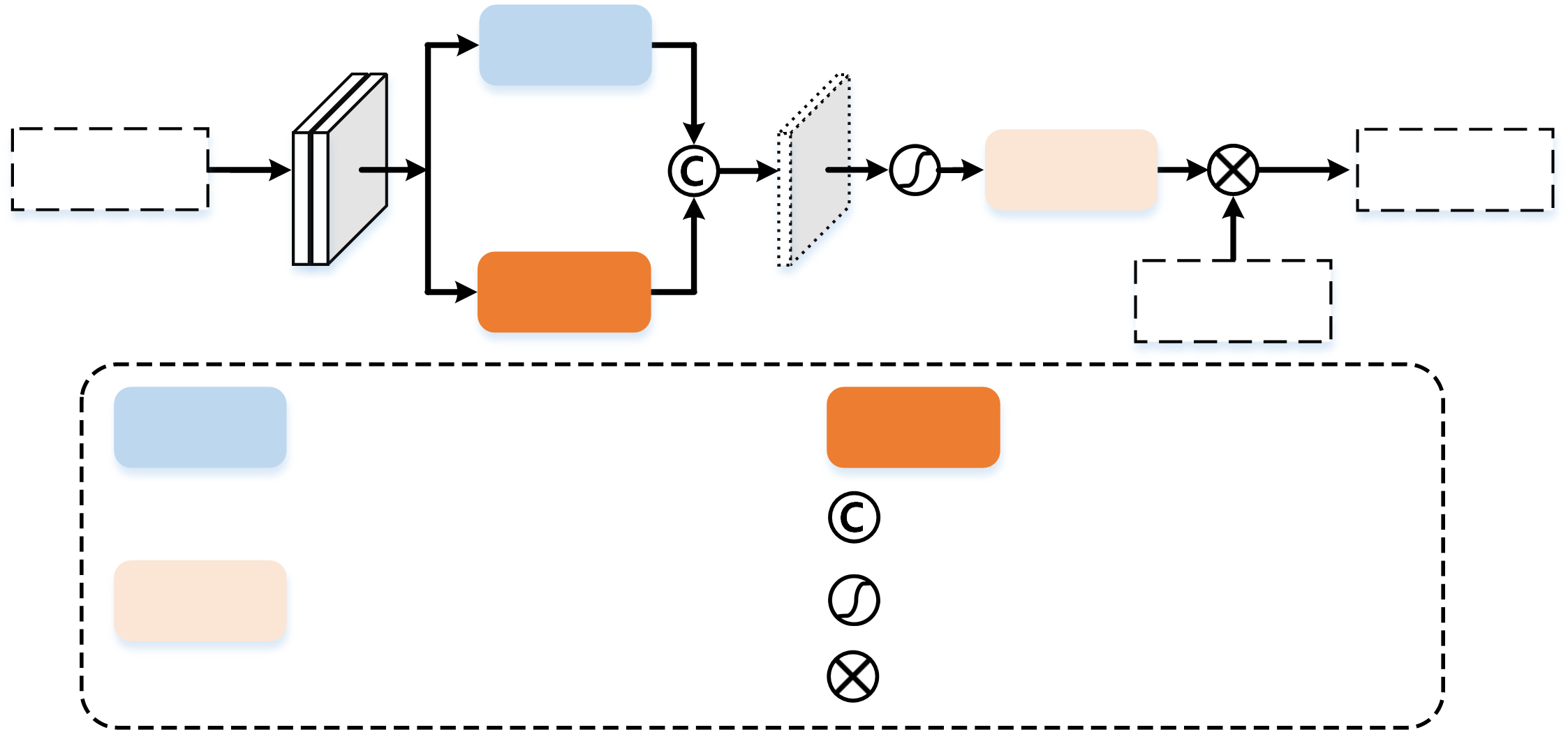} 
\put(3.9,35.4){\color{black}{$\bm{f}^{s}_{Ref}$}}
\put(50.2,43.9){\color{black}{$\bm{f}^{s}_{M}$}}

\put(72.4,40){\color{black}{$\bm{f}^{s}_{W}$}}
\put(75.7,26.9){\color{black}{$\bm{f}^{s}_{\emph{Dist}}$}}
\put(89.7,35.1){\color{black}{$\bm{f}^{s}_{\emph{Diff}}$}}

\put(32,42.9){\color{black}\scalebox{0.75}{$\mathsf{GAP(C)}$}}
\put(32,26.9){\color{black}\scalebox{0.75}{$\mathsf{GMP(C)}$}}
\put(64.7,34.7){\color{black}\scalebox{0.75}{$\mathsf{LAP(S)}$}}

\put(9,18.6){\color{black}\scalebox{0.75}{$\mathsf{GAP(C)}$}}
\put(54.4,18.6){\color{black}\scalebox{0.75}{$\mathsf{GMP(C)}$}}
\put(9,7.4){\color{black}\scalebox{0.75}{$\mathsf{LAP(S)}$}}

\put(24.3,19.4){\color{black}\scalebox{0.75}{$\bm {\mathsf{Channel \textbf{-} wise}}$}}
\put(20.2,17.4){\color{black}\scalebox{0.75}{$\bm {\mathsf{Global \, Average \, Pooling}}$}}
\put(68.7,19.4){\color{black}\scalebox{0.75}{$\bm {\mathsf{Channel \textbf{-} wise}}$}}
\put(65.9,17.4){\color{black}\scalebox{0.75}{$\bm {\mathsf{Global \, Max \, Pooling}}$}}
\put(24.3,8.4){\color{black}\scalebox{0.75}{$\bm {\mathsf{Spatial \textbf{-} wise}}$}}
\put(20.2,6.4){\color{black}\scalebox{0.75}{$\bm {\mathsf{Local \, Average \, Pooling}}$}}
\put(65.2,12.9){\color{black}\scalebox{0.75}{$\bm {\mathsf{Concatenate}}$}}
\put(67.6,7.6){\color{black}\scalebox{0.75}{$\bm {\mathsf{Sigmoid}}$}}
\put(61.6,2.5){\color{black}\scalebox{0.75}{$\bm {\mathsf{Element \textbf{-} wise Product }}$}}

\end{overpic}
\end{center}}
\vspace{-4.5ex}
\caption{\small
{Spatial attention for difference map calibration, where spatial attention based on reference feature is used to reweight distance map for generating calibrated difference map.}
}
\vspace*{-4mm}
\label{fig:SP}
\end{figure}
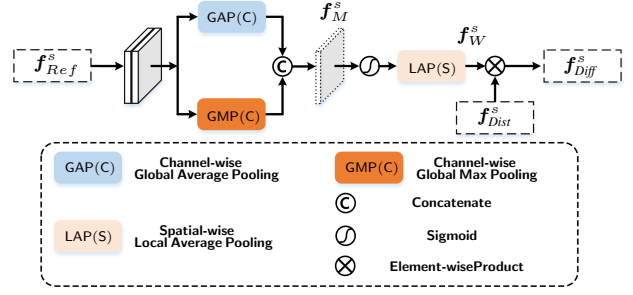
\textbf{Spatial Attention for Difference Map Calibration.}~Obviously, the contribution of image region to visual quality is spatially varying.
Informative regions have more influences and should be emphasized more when predicting the final score. %
In learning-based FR-IQA, ASNA~\cite{ASNA} computes spatial and channel attention based on decoder feature to improve MOS estimation.
%
%
%
Actually, the importance of local region should be determined by the reference image instead of decoder feature and distance map.
Thus, we adopt a much simple design by computing spatial attention based on reference feature while applying it on distance map to generate calibrated difference map.
%
%
%
As show in Fig.~\ref{fig:SP}, the spatial attention module takes reference feature $\bm{f}^{s}_{Ref}$ at scale $s$ as input.
Then, we use two $3 \times 3$ convolutional layers followed by global average pooling and max pooling along the channel dimension to form a feature map $\bm{f}^{s}_{M}$.
%
%
%
Finally, a $1 \times 1$ convolutional layer followed by sigmoid activation and local average pooling is deployed to generate spatial weighting map $\bm{f}^{s}_{W} \in \mathbb{R}^{ \frac{H}{p} \times \frac{W}{p}}$, where the size of the local average pooling region is set to $p \times p$.
Calibrated difference map $\bm{{f}}^{s}_{\emph{Diff}}$ can then be obtained by using $\bm{f}^{s}_{W}$ for reweighting each channel of distance map $\bm{{f}}^{s}_{\emph{Dist}}$ in an element-wise manner, while final score can be predicted by feeding $\bm{{f}}^{s}_{\emph{Diff}}$ into score prediction network.
%
%

%

\subsection{Network Structure of Binary Classifier}
The network structure of binary classifier is relatively simple, and contains two parts.
The first part involves the first 12 convolutional layers in VGG16 (\ie, 3 scales).
The second part has the same structure as the score prediction network in our FR-IQA model.

\section{Experiments}
\label{sec:exper}

In this section, we first introduce experiment settings and implementation details of the proposed method.
Then, we conduct ablation studies to analyze the proposed method, and compare it with state-of-the-art IQA methods on five benchmark datasets.
Finally, we evaluate the generalization ability of our method.

\begin{table}[!t]\scriptsize
\renewcommand\tabcolsep{2.0pt}
\centering
\caption{ Summary of five IQA databases, \ie, LIVE~\cite{LIVE}, CSIQ~\cite{CSIQ}, TID2013~\cite{TID2013}, KADID-10k~\cite{KADID} and PIPAL~\cite{PIPAL}. DMOS is inversely proportional to MOS.}
\vspace{-2mm}
\begin{tabular}{ccccccc}

\toprule[1pt]
Dataset&
\#Ref.&
\#Dis.&
\#Dis. Type&
\#Rating&
Rating Type&
Score Range\\
\bottomrule[1pt]

LIVE~\cite{LIVE}&
29&
779&
5&
25k&
DMOS&
[0, 100]\\

CSIQ~\cite{CSIQ}&
30&
866&
6&
5k&
DMOS&
[0, 1]\\

TID2013~\cite{TID2013}&
25&
3,000&
24&
524k&
MOS&
[0, 9]\\

KADID-10k~\cite{KADID}&
81&
10,125&
25&
30.4k&
MOS&
[1, 5]\\

PIPAL~\cite{PIPAL}&
250&
25,850&
40&
1.13m&
MOS&
[917, 1836]\\

\bottomrule[1pt]

\end{tabular}

\label{tab:IQA_data}
\vspace{-0.145in}

\end{table}

\subsection{Experiment Settings}
%
\textbf{Labeled Data.}
Five IQA datasets are employed in the experiments, including LIVE~\cite{LIVE}, CSIQ~\cite{CSIQ}, TID2013~\cite{TID2013}, KADID-10k~\cite{KADID} and PIPAL~\cite{PIPAL}, whose configurations are presented in Table \ref{tab:IQA_data}.
LIVE~\cite{LIVE}, CSIQ~\cite{CSIQ} and TID2013~\cite{TID2013} are three relatively small-scale IQA datasets, where distorted images only contain traditional distortion types (\eg, noise, downsampling, JPEG compression, \etc).
KADID-10k~\cite{KADID} further incorporates the recovered results of a denoising algorithm into the distorted images, resulting in a medium-sized IQA dataset.
%
%
Since the explicit splits of training, validation and testing are not given on these four datasets, we randomly partition the dataset into training, validation and testing sets by splitting reference images with ratios $60\%$, $20\%$, $20\%$, respectively.
%
To reduce the bias caused by a random split, we run the random splits ten times. %
On these four datasets, the comparison results are reported as the average of ten times evaluation experiments.

PIPAL~\cite{PIPAL} is a large-scale IQA dataset.
The training set consists of $200$ reference images and $23,200$ distorted images with resolution of $288 \times 288$.
The validation set consists of $25$ reference images and $1,000$ distorted images.
Since the testing set of PIPAL is not publicly available, we in this paper report the evaluation results on validation set via the online server\footnote{\scriptsize \url{{https://competitions.codalab.org/competitions/28050}}}.
The distorted images in PIPAL dataset include traditional distorted images and images restored by multiple types of image restoration algorithms (\eg, denoising, super-resolution, deblocking, etc.) as well as GAN-based restoration models.
%
%
%
It is worth noting that the distortion types in PIPAL validation set are unseen in the training set.
%

\textbf{Unlabeled Data.}
We take $1,000$ image patches ($288\times288$) randomly from DIV2K~\cite{DIV2K} validation set and Flickr2K~\cite{Flickr2K} as reference images in unlabeled data.
For the acquisition of distorted images, we adopt the following three manners:
(i) ESRGAN Synthesis: All the reference images are downsampled, and then super-resolved using 50 groups of intermediate ESRGAN models.
The restored images are regarded as distorted images in unlabeled data.
(ii) DnCNN Synthesis: We add Gaussian noises to reference images to obtain degraded images, which are restored using 50 groups of intermediate DnCNN models.
%
(iii) KADID-10k Synthesis: Following \cite{KADID}, we add 25 degradation types to reference images by randomly select 2 of 5 distortion levels for obtaining distortion images in unlabeled data.
More details of intermediate models of ESRGAN and DnCNN can be found in the supplementary material.
We note that ESRGAN and DnCNN are not adopted in validation set of PIPAL, guaranteeing non-intersection of distortion types in PIPAL validation set and our collected unlabeled data.
%
%

\textbf{Evaluation Criteria.}
Two evaluation criteria are reported for each experimental setup, \ie, Spearman Rank Correlation Coefficient (SRCC) for measuring prediction accuracy, and Pearson Linear Correlation Coefficient (PLCC) for measuring prediction monotonicity.
%

\subsection{Implementation Details}

We use the Adam optimizer~\cite{Adam} for all models presented in this paper with a batchsize of 32.
We randomly crop the image patches with size $224 \times 224$, and perform flipping (horizontal/vertical) and rotating ($90^\circ$, $180^\circ$, or $270^\circ$) on training samples for data augmentation.

\textbf{Supervised Learning.}
We train the proposed FR-IQA model with labeled data for total 20,000 iterations.
%
%
The learning rate is initialized to $1e\text{-}4$, and decreased to $1e\text{-}5$ after 10,000 iteration.
Moreover, we have found empirically that even if the training iterations are further increased, the IQA model will not get any performance improvement.

\textbf{Joint Semi-supervised and PU Learning.} We initialize the network parameters using the pre-trained IQA model with the learning rate of $1e\text{-}5$ for 20,000 iterations.
%
%
%
The pseudo MOS $y_{j}^{*}$ is initialized with the pre-trained IQA model.
Hyper-parameter $p$, \ie, the region size in local Sliced Wasserstein distance (LocalSW), is set to $8$ and $2$ for PIPAL and traditional IQA datasets, respectively.
%
The momentum parameter $\alpha$ is set to $0.95$.
Hyperparameter $\tau$ changes with iterations, \ie, $\tau=\max \{\tau_{0}^{t / T_{0}}, \tau_{\min }\}$ for $t$-th iteration, where parameters $\tau_{0}$, $T_{0}$ and $\tau_{\min }$ are set as $0.9$, $1,000$ and $0.5$, respectively.

\subsection{Ablation Study}
All the ablation experiments are performed on PIPAL~\cite{PIPAL} and KADID-10k~\cite{KADID}, considering that the distortion types of these two datasets are very different.

\textbf{Network Structure.}
\begin{table}[t] \scriptsize
\renewcommand\tabcolsep{6.0pt}
\centering
\caption{PLCC / SRCC performance with ablation studies about network structure performed on the PIPAL~\cite{PIPAL} and KADID-10k~\cite{KADID}.}
\label{tab:ab_arch}
\vspace{-3mm}
\begin{tabular}{c c c c c c}
\toprule
\multirow{2}[2]{*}{NO.}&\multirow{2}[2]{*}{DAB}&\multirow{2}[2]{*}{SA}&\multirow{2}[2]{*}{LocalSW}&PIPAL&KADID-10k\\
&&&&PLCC / SRCC&PLCC / SRCC\\ \midrule
1&\XSolidBrush&\XSolidBrush&\XSolidBrush&0.835 / 0.824&0.899 / 0.889\\
2&\Checkmark&\XSolidBrush&\XSolidBrush&0.843 / 0.837&0.908 / 0.905\\
3&\XSolidBrush&\Checkmark&\XSolidBrush&0.849 / 0.838&0.927 / 0.919\\
4&\Checkmark&\Checkmark&\XSolidBrush&0.852 / 0.849&0.941 / 0.940\\
5&\Checkmark&\XSolidBrush&\Checkmark&0.861 / 0.857&0.929 /0.925\\
6&\Checkmark&\Checkmark&\Checkmark&$\textbf{0.868}$ / $\textbf{0.868}$&$\textbf{0.943}$ / $\textbf{0.944}$\\
 \bottomrule
\end{tabular}
\vspace{-5mm}
\end{table}
We first study the effects of our three architectural components, \ie, Dual Attention Block (DAB), Spatial Attention (SA), and Local Sliced Wasserstein Distance (LocalSW).
In Table~\ref{tab:ab_arch}, one can see that on PIPAL dataset, removing the LocalSW results in the greatest performance degradation, which is mainly due to the additional computational error introduced by the spatial misalignment in the GAN-based distorted images.
When the SA module is eliminated, the IQA model assigns the same weight to different information content areas, resulting in low accuracy.
Similarly, DAB also contributes to the final performance.

\textbf{Training Strategy.}
We conduct ablation experiments on three different types of unlabeled data, \ie, ESRGAN Synthesis, DnCNN Synthesis, KADID-10k Synthesis, and compare the proposed JSPL with semi-supervised learning (SSL), \ie, combining labeled and unlabeled data without PU learning.
From Table~\ref{tab:ab learning}, we have the following observations:
(i) First, compared to the other two syntheses types, the distribution of unlabeled data using ESRGAN Synthesis is more consistent with the labeled PIPAL dataset, leading to the greater performance gains.
Similarly, the KADID-10k dataset has same distortion types with KADID-10k Synthesis.
%
It indicates that the inconsistent distribution between labeled and unlabeled data is a key issue for semi-supervised learning.
Therefore, in the subsequent experiments, we choose unlabeled data that are closer to the distribution of the labeled data.
(ii) Second, from the six sets of comparative experiments on SSL and JSPL, we can see that JSPL performs better than SSL.
This is because our JSPL can exclude negative outliers, making the distribution of labeled data and positive unlabeled data be more consistent, while SSL is adversely affected by these outliers.

%

\begin{table}[t]\scriptsize
\renewcommand\tabcolsep{1.0pt}
\begin{center}
\caption{PLCC / SRCC results obtained using different data settings with SL, SSL or JSPL manners on PIPAL~\cite{PIPAL} and KADID-10k~\cite{KADID}. }
\label{tab:ab learning}
\vspace{-3mm}
\begin{tabular}{c|ccc|ccc}
\hline
\multicolumn{1}{c|}{\multirow{2}{*}{Methods}} & \multicolumn{3}{c|}{PIPAL} & \multicolumn{3}{c}{KADID10k} \\
\multicolumn{1}{c|}{}& \multicolumn{2}{c}{Unlabeled Data} & PLCC / SRCC & \multicolumn{2}{c}{Unlabeled Data}        &     PLCC / SRCC                              \\  \hline
SL & \multicolumn{2}{c}{ - }& 0.868 / 0.868  & \multicolumn{2}{c}{ -}  & 0.943 / 0.944 \\ \hline
\multirow{3}{*}{SSL}   & \multicolumn{2}{c}{ ESRGAN Synthesis}  & 0.872 / 0.870 & \multicolumn{2}{c}{ ESRGAN Synthesis}& 0.930 / 0.932 \\
&\multicolumn{2}{c}{ DnCNN Synthesis}& 0.870 / 0.868 & \multicolumn{2}{c}{ DnCNN Synthesis}  & 0.945 / 0.944 \\
&\multicolumn{2}{c}{ KADID-10k Synthesis}      & 0.867 / 0.866       & \multicolumn{2}{c}{ KADID-10k Synthesis }   & 0.959 / 0.958  \\ \hline
\multirow{3}{*}{JSPL}   & \multicolumn{2}{c}{ ESRGAN Synthesis}&\textbf{0.877 / 0.874} & \multicolumn{2}{c}{ ESRGAN Synthesis}  & 0.945 / 0.948 \\
& \multicolumn{2}{c}{DnCNN Synthesis} & 0.875 / 0.872 & \multicolumn{2}{c}{ DnCNN Synthesis}  & 0.959 / 0.957\\
&\multicolumn{2}{c}{ KADID-10k Synthesis}  & 0.873 / 0.870&  \multicolumn{2}{c}{ KADID-10k Synthesis}& \textbf{0.963 / 0.961}\\
\hline
\end{tabular}
\vspace{-6mm}
\end{center}
\end{table}

\subsection{Comparison with State-of-the-arts}

\subsubsection{Evaluation on PIPAL Dataset}

\begingroup
\renewcommand{\arraystretch}{.9}
\begin{table}[t]
    \scriptsize
    \renewcommand\tabcolsep{3pt}
	\centering
	\caption{Performance comparison of IQA methods on PIPAL~\cite{PIPAL} dataset. 
	Some results are provided from the NTIRE 2021 IQA challenge report \cite{NTIRE}.}
\vspace{-3mm}
\begin{tabular}{lp{0.7cm}<{\centering}p{0.5cm}<{\centering}p{0.5cm}<{\centering}|l p{0.7cm}<{\centering}p{0.5cm}<{\centering}p{0.5cm}<{\centering}}
		\hline
		Methods& \hspace{-0.5cm}Category&PLCC &   SRCC&Methods &\hspace{-0.5cm}Category   &    PLCC &   SRCC     \\ \hline
        MA \cite{MA}&	\hspace{-0.5cm}\multirow{3}{*}{NR}     &   0.203   &   0.201   &PSNR                      &	\hspace{-0.5cm}\multirow{14}{*}{FR}     &  0.292   &   0.255   \\
        PI \cite{PI}	      &	     &   0.166   &   0.169   &SSIM \cite{SSIM}	        &        &  0.398   &   0.340     \\
        NIQE \cite{NIQE}	  &      &   0.102   &   0.064    &LPIPS-Alex \cite{LPIPS}	&	     &	0.646	&	0.628	\\\cline{1-4}
        VIF \cite{VIF}	      & \hspace{-0.5cm}\multirow{13}{*}{FR}     &   0.524   &   0.433   &LPIPS-VGG \cite{LPIPS}	&	     &	0.647	&	0.591	\\
        VSNR \cite{VSNR}	  &      &   0.375   &   0.321   &PieAPP \cite{PieAPP}	    &	     &	0.697	&	0.706	\\
        VSI \cite{VSI}	      &      &   0.516   &   0.450   &WaDIQaM-FR \cite{WaDIQaM}	&	     &	0.654	&	0.678\\
        MAD \cite{CSIQ}	      &      &   0.626   &   0.608   &DISTS \cite{DISTS}	    &	     &	0.686	&	0.674\\
        NQM \cite{NQM}	      &      &   0.416   &   0.346   &SWD \cite{PIPAL}          &	     &	0.668	&	0.661\\
        UQI \cite{UQI}	      &      &   0.548   &   0.486   &EGB~\cite{EGB}            &	     &  0.775   &   0.776\\
        IFC \cite{IFC}        &	     &   0.677   &   0.594   &DeepQA~\cite{DeepQA}      &	     &  0.795   &   0.785 \\
        GSM \cite{GSM}	      &      &   0.469   &   0.418   &ASNA~\cite{ASNA}          &	     &  0.831   &   0.824\\
        RFSIM \cite{RFSIM}    &	     &   0.304   &   0.266   &RADN~\cite{RADN}          &	     &  0.867   &   0.866\\
        SRSIM \cite{SRSIM}	  &      &   0.654   &   0.566   &IQMA~\cite{IQMA}          &	     &  0.876   &   0.872\\
        FSIM \cite{FSIM}      &	     &   0.561   &   0.467   &IQT~\cite{IQT}            &	     &  0.876   &   0.865\\ \cline{5-8}
        FSIMc \cite{FSIM}	  &      &   0.559   &   0.468   &Ours(SL)                  &	\hspace{-0.5cm}\multirow{2}{*}{FR}     &  0.868   &   0.868\\
        MS-SSIM \cite{MS-SSIM}&      &   0.563   &   0.486   &Ours(JSPL)                &	     &  $\textbf{0.877}$ & $\textbf{0.874}$\\\hline

	\end{tabular}
\label{tab:PIPAL}%
\vspace{-6mm}
\end{table}
\endgroup

As shown in Table~\ref{tab:PIPAL}, we compare 18 traditional evaluation metrics and 12 CNN-based FR-IQA models with the proposed model under two different learning strategies, \ie, supervised learning (SL) and JSPL.
Compared with traditional evaluation metrics, CNN-based FR-IQA models are proven to be more consistent with human subjective quality scoring.
Albeit retraining on the PIPAL dataset, the performance of pioneering CNN-based FR-IQA models, \eg, LPIPS~\cite{LPIPS}, WaDIQaM-FR~\cite{WaDIQaM} and DISTS \cite{DISTS} are still limited.
Although SWDN ~\cite{SWDN} designed a pixel-by-pixel alignment module to address the misalignment problem in GAN-based distortion, the corresponding feature extraction network is not sufficiently effective to achieve satisfactory result.
In contrast, considering both the properties of GAN-based distortion and the design of the feature extraction network, IQT~\cite{IQT}, IQMA~\cite{IQMA} and RADN~\cite{RADN} achieve top3 performance on PIPAL in published literatures.
%
Because of the spatial attention and the LocalSW module, the proposed method using supervised learning obtains superior performance than RADN~\cite{RADN} on PIPAL.
Although our FR-IQA model by adopting supervised learning strategy is slightly inferior to IQT~\cite{IQT} and IQMA~\cite{IQMA}, the proposed JSPL strategy significantly boosts its performance by exploiting adequate positive unlabled data while mitigating the adverse effects of outliers.
%
%
%
%
%

\subsubsection{Evaluation on Traditional Datasets}
\begin{table}\scriptsize
\renewcommand\tabcolsep{1.6pt}
  \centering
  \caption{Performance evaluation on the LIVE~\cite{LIVE}, CSIQ~\cite{CSIQ}, TID2013~\cite{TID2013} and KADID-10k~\cite{KADID} databases.}

\begin{tabular}{lc|cc|cc|cc|cc}
\hline
\multirow{2}[4]{*}{Methods}& \hspace{-0.5cm}\multirow{2}{*}{Category}& \multicolumn{2}{c|}{LIVE} & \multicolumn{2}{c|}{CSIQ} & \multicolumn{2}{c|}{TID2013}& \multicolumn{2}{c}{KADID-10k} \bigstrut\\
\cline{3-10}     & & SRCC  & PLCC  & SRCC  & PLCC  & SRCC  & PLCC & SRCC  & PLCC\bigstrut\\
\hline
BRISQUE~\cite{BRISQUE} &\hspace{-0.2cm}\multirow{14}{*}{NR} & 0.939 & 0.935 & 0.746 & 0.829 & 0.604 & 0.694&   -    & - \\
FRIQUEE~\cite{FRIQUEE}  & & 0.940  & 0.944 & 0.835 & 0.874 & 0.680  & 0.753&   -    & - \\
CORNIA~\cite{CORNIA} &  & 0.947 & 0.950  & 0.678 & 0.776 & 0.678 & 0.768&   0.541    & 0.580 \\
M3~\cite{M3}    & & 0.951 & 0.950  & 0.795 & 0.839 & 0.689 & 0.771&   -    & - \\
HOSA~\cite{HOSA}  & & 0.946 & 0.947 & 0.741 & 0.823 & 0.735 & 0.815 &   0.609    & 0.653\\
Le-CNN~\cite{Le-CNN}  & & 0.956 & 0.953 &   -    &   -    &   -    & - &   -    & -\\
BIECON~\cite{BIECON}  & & 0.961 & 0.962 & 0.815 & 0.823 & 0.717 & 0.762 &   -    & -\\
DIQaM-NR~\cite{WaDIQaM}  & & 0.960  & 0.972 &  -     &     -  & 0.835 & 0.855&   -    & - \\
WaDIQaM-NR~\cite{WaDIQaM} & & 0.954 & 0.963 & -      &  -     & 0.761 & 0.787&   -    & - \\
ResNet-ft~\cite{ResNet-ft}  & & 0.950  & 0.954 & 0.876 & 0.905 & 0.712 & 0.756&   -    & - \\
IW-CNN~\cite{ResNet-ft}  & & 0.963 & 0.964 & 0.812 & 0.791 & 0.800   & 0.802 &   -    & -\\
DB-CNN~\cite{DB-CNN} & & 0.968& 0.971& 0.946& 0.959 & 0.816 & 0.865&   0.501    & 0.569 \\
CaHDC~\cite{CaHDC} & & 0.965 & 0.964 & 0.903 & 0.914 & 0.862& 0.878&   -    & - \\
HyperIQA~\cite{hyperIQA} & & 0.962 & 0.966 & 0.923 & 0.942 &  0.729     & 0.775&   -    & -\\  \hline
PSNR        & \hspace{-0.2cm}\multirow{13}{*}{FR}& 0.873 & 0.865  & 0.810 & 0.819 & 0.687 & 0.677 & 0.676 & 0.675 \\
SSIM~\cite{SSIM} & & 0.948 & 0.937 & 0.865 & 0.852 & 0.727 & 0.777 & 0.724 & 0.717 \\
MS-SSIM \cite{MS-SSIM} & & 0.951 & 0.940 & 0.906 & 0.889 & 0.786 & 0.830  &0.826  &0.820  \\
VSI \cite{VSI} & & 0.952 & 0.948 & 0.942 & 0.928 & 0.897 & 0.900 & 0.879 & 0.877 \\
FSIMc \cite{FSIM} & & 0.965 & 0.961 & 0.931 & 0.919 & 0.851 & 0.877 & 0.854 & 0.850\\
MAD \cite{CSIQ}& & 0.967 & 0.968 & 0.947 & 0.950 & 0.781 & 0.827 & 0.799 & 0.799\\
VIF \cite{VIF}& & 0.964 & 0.960 & 0.911 & 0.913 & 0.677 & 0.771 & 0.679 & 0.687 \\
DeepSim~\cite{gao2017deepsim}& & 0.974  & 0.968 &  -     &     -  & 0.846 & 0.872&   -    & - \\
DIQaM-FR~\cite{WaDIQaM} &  & 0.966  & 0.977 &  -     &     -  & 0.859 & 0.880&   -    & - \\
WaDIQaM-FR~\cite{WaDIQaM} & & 0.970 & 0.980 & -      &  -     & $\textbf{0.940}$ & 0.946&   -    & - \\
DISTS \cite{DISTS} & &0.955  &0.955 &0.946 & 0.946&0.830 & 0.855& 0.887& 0.886\\
PieAPP \cite{PieAPP}&  &0.918  &0.909 &0.890 & 0.873&0.670 & 0.749& 0.836& 0.836\\
LPIPS \cite{LPIPS}&  & 0.932& 0.934&0.903 & 0.927&0.670 & 0.749& 0.843& 0.839\\
\hline
Ours(SL)   & \hspace{-0.2cm}\multirow{2}{*}{FR}&0.970&0.978&0.965&0.968&0.924&0.912&0.944&0.943\\
Ours(JSPL) & &$\textbf{0.980}$&$\textbf{0.983}$&$\textbf{0.977}$&$\textbf{0.970}$&$\textbf{0.940}$&$\textbf{0.949}$&$\textbf{0.961}$&$\textbf{0.963}$\\
\hline
\end{tabular}%

  \label{tab:trad_data}%
  \vspace*{-4mm}
\end{table}

Our methods with two learning manners, \ie, SL and JSPL, are compared with the competitors on the other four traditional IQA datasets, including LIVE~\cite{LIVE}, CSIQ~\cite{CSIQ}, TID2013~\cite{TID2013} and KADID-10k~\cite{KADID}.
%
From Table~\ref{tab:trad_data} we can observe that the FR-IQA models achieve a higher performance compared to the NR-IQA models, since the pristine-quality reference image provides more accurate reference information for quality assessment.
Although WaDIQaM-FR \cite{WaDIQaM} achieves almost the same performance with our method in terms of the SRCC metric on TID2013 dataset, but is inferior to ours on LIVE and PIPAL datasets, indicating its limited generalization ability.
%
On all testing sets, the proposed FR-IQA model with SL strategy still delivers superior performance, which reveals the effectiveness of the proposed spatial attention and LocalSW module.
By adopting JSPL strategy, our FR-IQA model achieves the best performance on all the four datasets.
%
%
More comparisons on individual distortion types and cross-datasets are provided in supplementary material.

\subsection{Evaluating Generalization Ability}

\begin{table}[!t]\scriptsize
\renewcommand\tabcolsep{6.0pt}
\begin{center}
\caption{PLCC / SRCC assessment about IQA models trained on different settings, and tested on the PIPAL~\cite{PIPAL} Val. }
\label{tab:test_gen}
\vspace{-3mm}

\begin{tabular}{c c c c}
\toprule

\multirow{2}[2]{*}{Methods} & Training Data&  PIPAL Val.\\
  & Labeld Data (\& Unlabeled Data)&  PLCC / SRCC\\ \midrule
  IQT(SL)&  PIPAL & 0.876 / 0.865\\
IQT(SL)&  KADID-10k & 0.741 / 0.718\\
IQT(SSL)&  KADID-10k \& ESRGAN Synthesis& 0.700 / 0.662 \\
IQT(JSPL)&  KADID-10k \& ESRGAN Synthesis & 0.794 / 0.783\\\midrule
Our(SL)&  PIPAL & 0.868 / 0.868\\
Ours(SL)&  KADID-10k & 0.756 / 0.770\\
Ours(SSL)&  KADID-10k \& ESRGAN Synthesis & 0.733 / 0.766 \\
Ours(JSPL)&  KADID-10k \& ESRGAN Synthesis & 0.804 / 0.801\\
 \bottomrule
\end{tabular}
\vspace{-8mm}
\end{center}
\end{table}

Considering that distortion types in KADID-10k and PIPAL are not similar, we adopt these two datasets for evaluating generalization ability of our method as well as IQT \cite{IQT}, a state-of-the-art method in Table \ref{tab:PIPAL}.
As shown in Table \ref{tab:test_gen}, both IQT and our method can obtain satisfying performance when keeping consistent validation and training sets from PIPAL.
However, significant performance degradations can be observed when applying the models learned based on KADID-10k to validation set of PIPAL.
This is because the distribution discrepancy between KADID-10k and PIPAL is severe, which cannot be addressed by SL strategy. %
By adopting SSL and JSPL, unlabeled data using ESRGAN Synthesis is introduced.
%
Although SSL utilizes unlabeled data, the performance drops can be observed for IQT and our method due to the effect of outliers, which demonstrates that the elimination of outliers is essential.
In contrast, our JSPL can exclude negative outliers while exploiting positive unlabeled data, significantly boosting generalization ability of IQT and our method.
In comparison to IQT with JSPL, our method with JSPL has better generalization ability, which can be attributed to the novel modules SA and LocalSW in our FR-IQA model.


\section{Conclusion}
In this paper, we proposed a joint semi-supervised and PU learning (JSPL) to exploit unlabelled data for boosting performance of FR-IQA, while mitigating the adverse effects of outliers.
We also introduced a novel FR-IQA network, embedding spatial attention and local sliced Wasserstein distance (LocalSW) for emphasizing informative regions and suppressing the effect of misalignment between distorted and pristine images, respectively.
Extensive experimental results show that the proposed JSPL algorithm can improve the performance of the FR-IQA model as well as the generalization capability.
In the future, the proposed JSPL algorithm can be extended to more challenging image quality assessment tasks, \eg, NR-IQA.

\section*{Acknowledgement}
This work was supported in part by National Key R\&D Program of China under Grant 2021ZD0112100, and National Natural Science Foundation of China under Grants No. 62172127, No. U19A2073 and No. 62102059.

{\small
\bibliographystyle{ieee_fullname}
\bibliography{egbib}
}

\twocolumn[
\begin{center}
	{\LARGE \textbf{Incorporating Semi-Supervised and Positive-Unlabeled Learning for Boosting
Full Reference Image Quality Assessment Supplemental Materials\\~\\~\\}}
\end{center}]

\renewcommand\thesection{\Alph{section}}
\renewcommand\thefigure{\Alph{figure}}
\renewcommand\thetable{\Alph{table}}
\renewcommand\thesubsection{\thesection.\arabic{subsection}}

\newcommand*{\affaddr}[1]{#1} 
\newcommand*{\affmark}[1][*]{\textsuperscript{#1}}
\newcommand*{\email}[1]{\texttt{#1}}

\newcommand{\tabincell}[2]{\begin{tabular}{@{}#1@{}}#2\end{tabular}}

\setcounter{section}{0}
\setcounter{table}{0}
\setcounter{figure}{0}

\hspace{-5mm}The content of this supplementary material includes:
\vspace{2mm}

    \hspace{-4mm}A. Limitation and Negative Impact in Sec.~\ref{sec:sm_LN}.

	\hspace{-4mm}B. ESRGAN and DnCNN Synthesis Process in Sec.~\ref{sec:sm_unlabel}.

    \hspace{-4mm}C. More Comparisons on Individual Distortion Types and Cross-dataset in Sec.~\ref{sec:sm_com}.

	\hspace{-4mm}D. More Ablation Studies in Sec.~\ref{sec:sm_ab}.

    \hspace{-4mm}E. Discussion in Sec.~\ref{sec:sm_BC}

	\hspace{-4mm}F. More Details on IQA Datasets in Sec.~\ref{sec:sm_data}.

\vspace{2mm}

\section{Limitation and Negative Impact}
\label{sec:sm_LN}
The proposed FR-IQA model predicts image quality by measuring the fidelity deviation from its pristine-quality reference.
Unfortunately, in the vast majority of practical applications, reference images are not always available or difficult to obtain, which indicates our method is limited especially for authentically-distorted images.

\section{ESRGAN and DnCNN Synthesis Process}
\label{sec:sm_unlabel}

For ESRGAN Synthesis, we adopt the DIV2K~\cite{DIV2K} training set as clean high-resolution (HR) images and employ the bicubic downsampler with the scale factor $2$ to obtain the low-resolution (LR) images.
Then, we retrain the original ESRGAN model using HR-LR pairs with the size of $128 \times 128$ and $64 \times 64$ cropped from the training HR and LR images, respectively.
The ESRGAN model is trained with the GAN loss for 50 epochs and 50 groups of intermediate ESRGAN models are obtained.
The learning rate is initialized to $2e\text{-}4$ and then decayed to $2e\text{-}5$ after 20 epochs.
We take $1,000$ image patches ($288\times288$) randomly from DIV2K~\cite{DIV2K} validation set and Flickr2K~\cite{Flickr2K} as reference images in unlabeled data, which are propagated into the bicubic downsampler to obtain the degraded images.
The corresponding distorted images can be obtained by feeding the degraded images into 50 groups of intermediate ESRGAN models.

For synthetic noises in DnCNN Synthesis, we use the additive white Gaussian noise with noise level $25$.
DnCNN is trained to learn a mapping from noisy image to denoising result.
The DnCNN model is trained with the MSE loss for 50 epochs and 50 groups of intermediate DnCNN models are obtained.
The learning rate is fixed to $1e\text{-}4$ and then decayed to $1e\text{-}5$ after 25 epochs.
Similarly, we also take same $1,000$ image patches as reference images in unlabeled data.
The restored images can be achieved by feeding the noisy images into 50 groups of intermediate DnCNN models, which are regared as the corresponding distorted images in unlabeled data.


\section{More Comparisons on Individual Distortion Types and Cross-dataset}
\label{sec:sm_com}

\begin{table}\scriptsize
\renewcommand\tabcolsep{8.0pt}
  \centering
  \caption{SRCC comparisons on individual distortion types on the LIVE database. {\color{red}Red} and {\color{blue}{blue}} {\color{black} are utilized to indicate top {\color{red}1\textsuperscript{st}} and {\color{blue}2\textsuperscript{nd}} rank, respectively}.}

\begin{tabular}{c|c c c c c}
\hline
Database & \multicolumn{5}{c}{LIVE} \\ \hline
Type & WN & JPEG  & JP2K & FF & GB    \\ \hline
WaDIQaM-FR~\cite{WaDIQaM} & 0.975 & 0.959  & 0.934 & 0.941 & 0.915    \\
DISTS \cite{DISTS} & 0.969 & 0.982  & \textcolor[rgb]{1,0,0}{0.971} & 0.961 & \textcolor[rgb]{1,0,0}{0.969}   \\
PieAPP \cite{PieAPP} &  0.963 & 0.941  & 0.885 & 0.920 & 0.867    \\
LPIPS \cite{LPIPS} & 0.968 & 0.982  & \textcolor[rgb]{0,0,1}{0.968} & 0.955 & 0.918  \\
\hline
our(SL)   & \textcolor[rgb]{0,0,1}{0.983} & \textcolor[rgb]{0,0,1}{0.984}  & 0.952 & \textcolor[rgb]{0,0,1}{0.967}& 0.912 \\
our(JSPL) & \textcolor[rgb]{1,0,0}{0.984} & \textcolor[rgb]{1,0,0}{0.986}  & 0.959 & \textcolor[rgb]{1,0,0}{0.968} & \textcolor[rgb]{0,0,1}{0.943} \\ \hline
\end{tabular}%

  \label{tab:trad_type}%
\end{table}%

\textbf{Comparisons on Individual Distortion Types.}
To further investigate the behaviors of our proposed method, we exhibit the performance on individual distortion type and compare it with several competing FR-IQA models on LIVE.
The LIVE dataset contains five distortion types, \ie, additive white Gaussian noise (WN), JPEG compression (JPEG), JPEG2000 compression (JP2K), Gaussian blur (GB) and Rayleigh fast-fading channel distortion (FF).
As shown in Table~\ref{tab:trad_type}, the average SRCC values of above ten groups are reported.
It is worth noting that our methods achieve significant performance improvements on three distortion types, \ie, WN, JPEG and FF.
Overall, better consistency with subjective scores and the consistently stable performance across different distortion types of the proposed scheme makes it the best IQA metric among all the compared metrics.

\begin{table}\scriptsize
\renewcommand\tabcolsep{1pt}
  \centering
  \caption{SRCC comparisons on different cross-dataset with the PIPAL as training set. {\color{red}Red} and {\color{blue}{blue}} {\color{black} are utilized to indicate top {\color{red}1\textsuperscript{st}} and {\color{blue}2\textsuperscript{nd}} rank, respectively}.}

\scalebox{0.95}{\begin{tabular}{c|c |c c c c}
\hline
\multirow{2}[2]{*}{Methods}&Traingning Set & \multicolumn{4}{c}{Test Sets} \\
 & Labeld Data (\& Unlabeled Data) & LIVE  & CSIQ & TID2013 & KADID-10k    \\ \hline
WaDIQaM-FR~\cite{WaDIQaM} & PIPAL & 0.910  & 0.877 & 0.802 & 0.713   \\
DISTS \cite{DISTS} & PIPAL & 0.913  & 0.876 & 0.803 & 0.706  \\
PieAPP \cite{PieAPP} & PIPAL & 0.904  & 0.875 & 0.762 & 0.699    \\
LPIPS \cite{LPIPS} & PIPAL & 0.908  & 0.863 & 0.795 & 0.717  \\
IQT \cite{IQT} & PIPAL & 0.917  & \textcolor[rgb]{0,0,1}{0.880} & 0.796 & \textcolor[rgb]{0,0,1}{0.718}  \\
\hline
our(SL)   & PIPAL & \textcolor[rgb]{0,0,1}{0.919}  & 0.873 & \textcolor[rgb]{0,0,1}{0.804} & 0.717 \\
our(JSPL) & PIPAL \&  KADID-10k Synthesis & \textcolor[rgb]{1,0,0}{0.930}  & \textcolor[rgb]{1,0,0}{0.894} & \textcolor[rgb]{1,0,0}{0.812} & \textcolor[rgb]{1,0,0}{0.776} \\ \hline
\end{tabular}}

  \label{tab:cross_pipal}%
\end{table}%

\textbf{Comparisons on Cross-dataset.}
To verify the generalization capability, we further evaluate the proposed method on three groups of cross-dataset settings.
We compare five FR-IQA methods, including: WaDIQaM-FR~\cite{WaDIQaM}, DISTS \cite{DISTS}, PieAPP \cite{PieAPP}, LPIPS \cite{LPIPS} and IQT \cite{IQT} with the proposed model under two different learning strategies, \ie, SL and JSPL.
We retrain the DISTS \cite{DISTS}, PieAPP \cite{PieAPP} and LPIPS \cite{LPIPS} by the source codes provided by the authors.
%
Although the source training code for WaDIQaM-FR and IQT is not publicly available, we reproduce WaDIQaM-FR~\cite{WaDIQaM} and IQT \cite{IQT}, and achieve the similar performance of the original paper.
From Table~\ref{tab:cross_pipal}, all FR-IQA models with supervised learning (SL) are trained using the largest human-rated IQA dataset, \ie, PIPAL, so the results on the other four test datasets are relatively close.
Because our approach with JSPL makes full use of unlabeled KADID-10k Synthesis which contains the same distortion types with KADID-10k, the higher performance on KADID-10k can be obtained.

\begin{table}\scriptsize
\renewcommand\tabcolsep{1pt}
  \centering
  \caption{SRCC comparisons on different cross-dataset with the KADID10k as training set. {\color{red}Red} and {\color{blue}{blue}} {\color{black} are utilized to indicate top {\color{red}1\textsuperscript{st}} and {\color{blue}2\textsuperscript{nd}} rank, respectively}.}

\scalebox{0.95}{\begin{tabular}{c|c |c c c c}
\hline
\multirow{2}[2]{*}{Methods}&Traingning Set & \multicolumn{4}{c}{Test Sets} \\
 & Labeld Data (\& Unlabeled Data) & LIVE  & CSIQ & TID2013 & PIPAL Val.    \\ \hline
WaDIQaM-FR~\cite{WaDIQaM} & KADID-10k & 0.948  & 0.931 & 0.861 & 0.712   \\
DISTS \cite{DISTS} & KADID-10k & 0.954  & 0.939 & 0.881 & 0.703  \\
PieAPP \cite{PieAPP} &  KADID-10k& 0.917  & 0.936 & 0.856 & 0.633    \\
LPIPS \cite{LPIPS} & KADID-10k & 0.932  & 0.917 & 0.821 & 0.671  \\
IQT \cite{IQT} & KADID-10k & 0.970  & 0.943 & 0.899 & 0.718  \\
\hline
our(SL)   & KADID-10k & \textcolor[rgb]{0,0,1}{0.973}  & \textcolor[rgb]{0,0,1}{0.951} & \textcolor[rgb]{0,0,1}{0.908} & \textcolor[rgb]{0,0,1}{0.770} \\
our(JSPL) & KADID-10k \&  KADID-10k Synthesis & \textcolor[rgb]{1,0,0}{0.974}  & \textcolor[rgb]{1,0,0}{0.953} & \textcolor[rgb]{1,0,0}{0.910} & - \\
our(JSPL) & KADID-10k \&  ESRGAN Synthesis & -  & - & - & \textcolor[rgb]{1,0,0}{0.801} \\ \hline
\end{tabular}}

  \label{tab:cross_kadid}%
\end{table}%

\begin{table}\scriptsize
\renewcommand\tabcolsep{1pt}
  \centering
  \caption{SRCC comparisons on different cross-dataset with the TID2013 as training set. {\color{red}Red} and {\color{blue}{blue}} {\color{black} are utilized to indicate top {\color{red}1\textsuperscript{st}} and {\color{blue}2\textsuperscript{nd}} rank, respectively}.}

\scalebox{0.95}{\begin{tabular}{c|c |c c c c}
\hline
\multirow{2}[2]{*}{Methods}&Traingning Set & \multicolumn{4}{c}{Test Sets} \\
 & Labeld Data (\& Unlabeled Data) & LIVE  & CSIQ & KADID-10k & PIPAL Val.    \\ \hline
WaDIQaM-FR~\cite{WaDIQaM} & TID2013 & 0.911  & 0.913 & 0.760 & 0.552   \\
DISTS \cite{DISTS} & TID2013 & 0.923  & 0.914 & 0.737 & 0.458  \\
PieAPP \cite{PieAPP} &  TID2013 & 0.888  & 0.886 & 0.573 & 0.401    \\
LPIPS \cite{LPIPS} & TID2013 & 0.895  & 0.913 & 0.761 & 0.595  \\
IQT \cite{IQT} & TID2013 & 0.940  & 0.929 &\textcolor[rgb]{0,0,1}{0.775} & 0.639  \\
\hline
our(SL)   & TID2013 & \textcolor[rgb]{0,0,1}{0.944}  & \textcolor[rgb]{0,0,1}{0.932} & 0.762& \textcolor[rgb]{0,0,1}{0.651} \\
our(JSPL) & TID2013 \&  KADID-10k Synthesis & \textcolor[rgb]{1,0,0}{0.948}  & \textcolor[rgb]{1,0,0}{0.934} & \textcolor[rgb]{1,0,0}{0.795} & - \\
our(JSPL) & TID2013 \&  ESRGAN Synthesis & -  & - & - & \textcolor[rgb]{1,0,0}{0.699} \\ \hline
\end{tabular}}

  \label{tab:cross_tid}%
\end{table}%

From Table.~\ref{tab:cross_kadid}, all FR-IQA models with supervised learning (SL) are trained on KADID-10k, which contains the most diverse traditional distortion types.
Therefore, compared to training on PIPAL or TID2013, all the FR-IQA methods achieve the best performance on traditional IQA datasets, \eg, LIVE and CSIQ.
Compared to other FR-IQA models, the proposed FR-IQA designs the spatial attention to deploy in computing difference map for emphasizing informative regions, and achieves the best performance in all FR-IQA models with supervised learning.
However, when testing on PIPAL which contains distortion images restored by multiple types of image restoration algorithms as well as GAN-based restoration, significant performance degradation can be observed due to the distribution variation among different datasets.
To alleviate this problem, the proposed JSPL strategy can improve performance to some extent for the use of unlabeled data.
%

From Table.~\ref{tab:cross_tid}, all FR-IQA models with supervised learning (SL) are trained on TID2013.
Due to fewer human-annotations and distorted samples are provided in TID2013, compared to KADID-10k, performance drop can be observed on traditional datasets, \eg, LIVE and CSIQ, which indicates the collection of massive MOS annotations is beneficial to the performance improvement.
However, the collection of massive MOS annotations is very time-consuming and cumbersome.
In this work, we consider a more encouraging and practically feasible SSL setting, \ie, training FR-IQA model using labeled data as well as unlabeled data.
Based on three groups of cross-dataset experiments, the proposed JSPL can exploit positive unlabeled  data, and significantly  boost the performance and  the generalization ability of FR-IQA.

\section{More Ablation Studies}
\label{sec:sm_ab}

\begin{table}\scriptsize
\renewcommand\tabcolsep{20.0pt}
\centering
\caption{PLCC / SRCC results for computing spatial attention based on different features.}
\label{tab:ab_sa}
\vspace{-3mm}

\begin{tabular}{c c}
\toprule
Based on &PIPAL Val.\\ \midrule
Reference feature $\bm{f}^{s}_{Ref}$&$\textbf{0.868}$ / $\textbf{0.868}$\\
Distortion feature $\bm{f}^{s}_{Dis}$&0.861 / 0.860\\
Distance map $\bm{{f}}^{s}_{\emph{Dist}}$& 0.864 / 0.864 \\
\bottomrule
\end{tabular}
\vspace{-3mm}
\end{table}

\begin{table}\scriptsize
\renewcommand\tabcolsep{20.0pt}
\centering
\caption{\footnotesize Performance on different attention mechanism on PIPAL.} \label{tab:saca}
    \scriptsize
    \vspace{-3mm}
    \scalebox{0.80}{
    \begin{tabular}{p{1.5cm}<{\centering} p{1.5cm}<{\centering} p{1.5cm}<{\centering}}
\toprule
\multicolumn{2}{c}{Attention Mechanism}&\multirow{2}[2]{*}{SRCC}\\
 Spatial&Channel&\\ \midrule
\XSolidBrush&\XSolidBrush&0.857\\
\Checkmark&\XSolidBrush&$\textbf{0.868}$\\
\XSolidBrush&\Checkmark&0.840\\
\Checkmark&\Checkmark&0.859\\ \bottomrule
\end{tabular}}
\vspace{-3mm}
\end{table}

\begin{table}\scriptsize
\renewcommand\tabcolsep{20.0pt}
\centering
\caption{PLCC / SRCC results for varying threshold parameter (\ie, $\tau_{min}$) on PIPAL~\cite{PIPAL} and KADID-10k~\cite{KADID}.}
\label{tab:ab_tau}
\vspace{-3mm}

\scriptsize
\begin{tabular}{c c c}
\toprule
\multirow{2}[2]{*}{$\tau_{min}$}&PIPAL &KADID-10k\\
&PLCC / SRCC&PLCC / SRCC\\ \midrule
$0.4$&0.872 / 0.870& 0.951 / 0.949\\
$0.5$&$\textbf{0.877}$ / $\textbf{0.874}$& $\textbf{0.963}$ / $\textbf{0.961}$ \\
$0.6$&0.874 / 0.872& 0.955 / 0.955\\
\bottomrule
\end{tabular}
\vspace{-3mm}
\end{table}

\begin{table}\scriptsize
\renewcommand\tabcolsep{20.0pt}
\centering
\caption{\footnotesize SRCC performance on different sliced Wasserstein. $p$ denotes local region size.} \label{tab:sw}
    \scriptsize
    \vspace{-3mm}
    \scriptsize
    \scalebox{0.83}{
 \begin{tabular}{p{0.4cm}<{\centering} p{1.0cm}<{\centering} p{0.6cm}<{\centering} p{1.4cm}<{\centering}}
\toprule
\multicolumn{2}{c}{Methods}&PIAPL&KADID-10k\\ \midrule
\multicolumn{2}{c}{Global}&0.755&0.509\\ \hline
\multirow{6}{*}{Local}&$p=32$&0.820&0.881\\
&$p=16$&0.862&0.928\\
&$p=8$&$\textbf{0.868}$&0.933\\
&$p=4$&0.866&0.939\\
&$p=2$&0.864&\textbf{0.944}\\
&$p=1$&0.857&0.940\\
\bottomrule
\end{tabular}}
\vspace{-3mm}
\label{tab:p}%
\end{table}

\textbf{Spatial Attention.}
As far as the design of spatial attention, we adopt a much simple design by computing spatial attention based on the reference feature while applying it to the distance map to generate calibrated difference map.
We conduct the ablation study by computing spatial attention based on different features, \ie, the reference feature $\bm{f}^{s}_{Ref}$, the distortion feature $\bm{f}^{s}_{Dis}$ and the distance map $\bm{{f}}^{s}_{\emph{Dist}}$.
%
%
Considering the superiority of extracting features from reference in Table~\ref{tab:ab_sa}, individual spatial attention on reference features is finally adopted in our method, while in ASNA~\cite{ASNA}, spatial attention and channel attention are directly adopted on distance map.
In Table~\ref{tab:saca}, ablation studies on attention mechanism are reported, where individual spatial attention on reference features performs best.
In IW-SSIM~\cite{IW-SSIM}, spatially local information is suggested as one key factor for assessing distortions, which motivates us to only adopt spatial attention.
%

\textbf{Hyper-parameter $\tau_{min}$.}
We study the effects of threshold parameter, \ie, $\tau_{min}$ on PIPAL~\cite{PIPAL} and KADID-10k~\cite{KADID}.
From Table~\ref{tab:ab_tau}, the best performance is achieved on both two datasets when $\tau_{min}$ is set to $0.5$.

\textbf{LocalSW.}
As for LocalSW, we suggest that local regions with proper size are more suitable for assessing distortions.
As shown in Table \ref{tab:p}, region size $p=8$ is the best choice on PIPAL, while original sliced Wasserstein (Global) yields significant performance drop.
We further study the effects of hyper-parameter $p$ on PIPAL~\cite{PIPAL} and KADID-10k~\cite{KADID}, because the distortion types of these two datasets are very different.
Due to the spatial misalignment properties of GAN-based distorted images in PIPAL, when the region size $p$ is set to $8$, the proposed LocalSW can compare the features within the most appropriate range around the corresponding position as shown in Table~\ref{tab:p}.
When applied to traditional dataset, \ie,  KADID-10k, the LocalSW with the hyper-parameter $p=2$ achieves the best results.

\begin{table}\scriptsize
\renewcommand\tabcolsep{12pt}
  \centering
  \caption{PLCC / SRCC comparisons on different FR-IQA with SL or JSPL training on PIPAL. {\color{red}Red} and {\color{blue}{blue}} {\color{black} are utilized to indicate top {\color{red}1\textsuperscript{st}} and {\color{blue}2\textsuperscript{nd}} rank, respectively}.}

\begin{tabular}{c c c}
\hline
 Method& SL  & JSPL \\ \hline
WaDIQaM-FR~\cite{WaDIQaM} & 0.778 / 0.761  & 0.793 / 0.775   \\
DISTS \cite{DISTS} & 0.813 / 0.806 & 0.822 / 0.812   \\
PieAPP \cite{PieAPP} &  0.785 /	0.778   & 0.806 / 0.796   \\
LPIPS \cite{LPIPS} & 0.790 / 0.790  & 0.809 / 	0.802\\
IQT \cite{IQT} & \textcolor[rgb]{1,0,0}{0.876} / \textcolor[rgb]{0,0,1}{0.865}  & \textcolor[rgb]{0,0,1}{0.876} / \textcolor[rgb]{0,0,1}{0.873}  \\
our & \textcolor[rgb]{0,0,1}{0.868} / \textcolor[rgb]{1,0,0}{0.868}  & \textcolor[rgb]{1,0,0}{0.877} / \textcolor[rgb]{1,0,0}{0.874} \\ \hline
\end{tabular}%
\label{tab:JSPL_FR}%
\end{table}%

\textbf{Applying JSPL to Different FR-IQA models.}
To verify the generalization capability of JSPL, we apply the proposed JSPL to 6 different FR-IQA models, and use the PIPAL training set to retrain the 6 different FR-IQA models.
From Table~\ref{tab:JSPL_FR}, the pioneering CNN-based FR-IQA models, \eg, WaDIQaM-FR~\cite{WaDIQaM}, DISTS \cite{DISTS}, PieAPP \cite{PieAPP} and LPIPS \cite{LPIPS} trained with PIPAL in supervised learning manner perform better than the original models (Table~\ref{tab:PIPAL} in the manuscript) on PIPAL validation set.
In terms of the SRCC metric, the proposed FR-IQA achieves the best performance with the help of LocalSW and spatial attention.
Compared to the supervised learning, the proposed JSPL can further boost the performance of all six FR-IQA models, which indicates that the proposed learning strategy has good generalization ability.

\section{Discussion}
\label{sec:sm_BC}

\begin{table}\scriptsize
\renewcommand\tabcolsep{12pt}
  \centering
  \caption{Total number of distortion images (\# U), number of positive samples (\# PU) and number of negative samples (\# NU) in the different distortion types.}
\begin{tabular}{c c c c}
\hline
Distortion Types& \# U  & \# PU & \# NU \\ \hline
DnCNN denoising algorithm & 2,000  & 1,996&  4\\
Gaussian blur & 2,000  & 1,996&  4\\
Additive white Gaussian noise & 2,000  & 1,979 & 21     \\
Color over-saturation & 2,000  & 0 & 2,000 \\
Color blocking & 2,000  &10 & 1,990   \\
Sharpness & 2,000  & 12 & 1,988  \\ \hline
\end{tabular}%
\label{tab:BC}%
\end{table}%

\begin{figure}[t]
\begin{center}
\scalebox{1.00}{
\begin{tabular}[t]{c@{ }c@{ }c@{ }c}
\includegraphics[width=.11\textwidth]{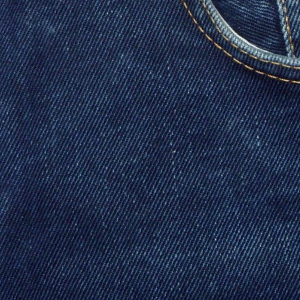}& \hspace{0.1mm}
\includegraphics[width=.11\textwidth]{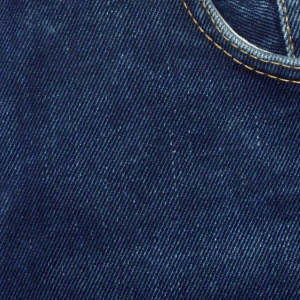}& \hspace{0.1mm}
\includegraphics[width=.11\textwidth]{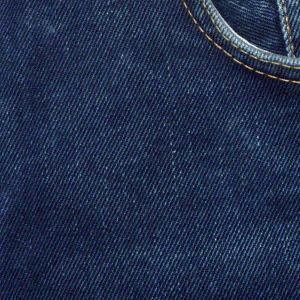}& \hspace{0.1mm}
\includegraphics[width=.11\textwidth]{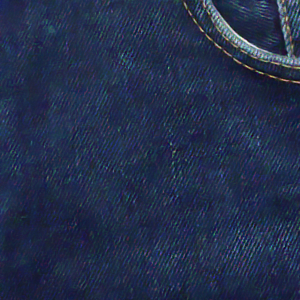}\\
\includegraphics[width=.11\textwidth]{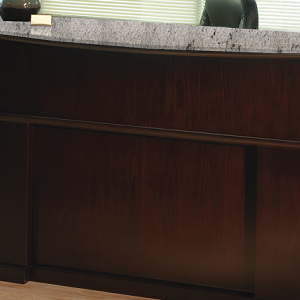}& \hspace{0.1mm}
\includegraphics[width=.11\textwidth]{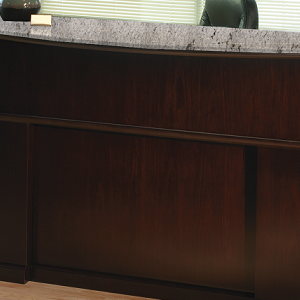}& \hspace{0.1mm}
\includegraphics[width=.11\textwidth]{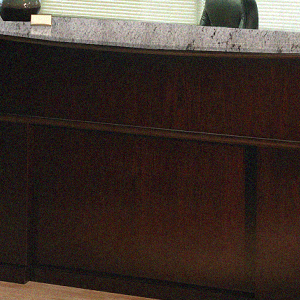}& \hspace{0.1mm}
\includegraphics[width=.11\textwidth]{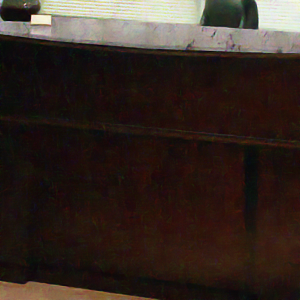}\\
(a) Reference & (b) GB & (c) WN & (d) DN

\end{tabular}
}
\end{center}
\vspace*{-2mm}
\caption{Visualization of the excluded outliers, \ie, the corresponding reference images, DnCNN denoising (DN) distorted images, Gaussian blur (GB) distorted images and additive white Gaussian noise (WN) distorted images.}
\label{fig:out}
\vspace{-2mm}
\end{figure}

\textbf{More Analysis on Binary Classifier.} The labeled IQA datasets~\cite{PIPAL, KADID} selected reference images which are representative of a wide variety of real-world textures, and should not be over-smooth or monochromatic.
The reference images in unlabeled data are chosen randomly from DIV2K~\cite{DIV2K} validation set and Flickr2K~\cite{Flickr2K}, hence a small number of images may not meet the requirements.
The unlabeled data may also contain distorted images which differ significantly from the distribution of the labeled data.

To verify that the binary classifier can eliminate the outliers mentioned above, we conduct the experiment to analyze the positive unlabeled data and outliers selected by the classifier.
Take our FR-IQA as an example, the PIPAL training samples are selected as labeled data and the unlabeled data are considered to use the KADID-10k Synthesis, which contain multiple distortion types and are more useful for analysis than ESRGAN Synthesis and DnCNN Synthesis.
We choose the 6 distortion types out of a total of 25 for analysis, \ie, DnCNN denoising algorithm, Gaussian blur, additive white Gaussian noise, color over-saturation, color blocking and sharpness.
As shown in Table~\ref{tab:BC}, each distortion type contains 2,000 distorted images.
The three types of distortion, \ie, DnCNN denoising algorithm, Gaussian blur and additive white Gaussian noise, are present on both PIPAL and KADID-10k Synthesis and are therefore heavily selected as positive unlabeled data by the classifier for semi-supervised learning of IQA models.
In contrast, the other three types of distortion are unseen for PIPAL, and the corresponding distortion images differ significantly from the distribution of the labeled data in PIPAL, which are excluded by the classifier.
Furthermore, we find that the 4 outliers in the DnCNN denoising algorithm or Gaussian blur settings are synthesized based on the same two reference images, as shown in Fig.~\ref{fig:out}.
We consider the reason is that those two reference images are over-smooth or monochromatic, which lack real-world textures and not meet the requirement for reference images.
%
%
%
In summary, the proposed JSPL is leveraged to identify negative samples from unlabeled data, \eg, reference images that lack real-world textures or distorted images that differ significantly from the labeled data.

\begin{table}[t] \scriptsize
\centering
\vspace{-3mm}
   \caption{\footnotesize SRCC comparison on different numbers of reference images and distortion types.}
\vspace{-3mm}
\scalebox{0.85}{
\begin{tabular}{lp{1.5cm}<{\centering} p{1.5cm}<{\centering} p{1.5cm}<{\centering}}
    \toprule
\diagbox [width=14em,trim=l] {Distortion}{\# Reference image} & 1,000 & 500 & 100 \\
\hline
Full 25 types & $\textbf{0.776}$ & 0.766 &  0.739 \\
10 types with top-10 ratios& 0.770 &   0.759&   0.735\\
10 types with bottom-10 ratios& 0.743 &   0.736&  0.719\\
    \bottomrule
\end{tabular}}
\label{tab:diss2}
 \vspace{-2mm}
\end{table}

\textbf{More discussion on how much unlabeled data and number of distortions.}
We use the PIPAL training set as labeled set, and use several representative distortion models to synthesize unlabeled samples.
Specifically, there are total 25 distortion types in KADID-10k and 1,000 reference images.
Based on the trained classifier, the ratios $\rho=\frac{\text{positive unlabeled samples}}{\text{outliers}}$ can be computed for 25 distortion types.
In Table~\ref{tab:BC}, distortion types with top-3 and bottom-3 ratios are presented.
Taking KADID-10k as testing bed, we discuss the sensitivity of our JSPL with different numbers of unlabeled samples and distortion types.
As for the number of reference images, we set it as 1,000, 500 and 100.
As for distortions, we adopt three settings, \ie, full 25 types, 10 types with top-10 $\rho$ ratios and 10 types with bottom-10 $\rho$ ratios.
%
The results are summarized in Table \ref{tab:diss2}.
%
We can observe that:
(i) Benefiting from unlabeled samples, our JSPL contributes to performance gains for any setting, \ie, the models in Table \ref{tab:diss2} are all superior to the model trained on only labeled data (SRCC = 0.717 by Our(SL) in Table \ref{tab:cross_pipal}).
(ii) When reducing the number of reference images from 1,000 to 500, our JSPL slightly degrades for all the three distortion settings.
And it is reasonable that the performance of JSPL is close to Our(SL) when few unlabeled samples are exploited.
(iii) As for distortions, the IQA models with bottom-10 $\rho$ ratios are notably inferior to Our(JSPL), indicating that JSPL can well exclude outliers.

\section{More Details on IQA Datasets}
\label{sec:sm_data}

Details of the different IQA datasets containing the distortion types can be viewed in Table~\ref{tab:list}.
Among them, the KADID-10k contains the richest traditional distortion types and the PIAPL contains the richest distortion types of the recovery results.

\begin{figure*}\scriptsize
\begin{center}

\begin{tabular}{c@{ } c@{ }  c@{ } c@{ }  c@{ } c@{ } c@{ }}

    \includegraphics[width=.140\textwidth]{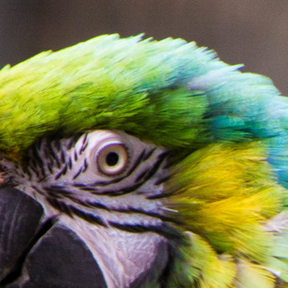}&
    \includegraphics[width=.140\textwidth]{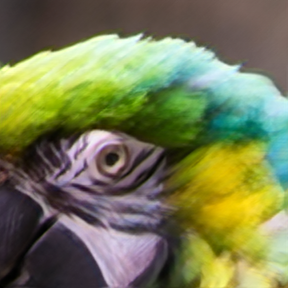}&
    \includegraphics[width=.140\textwidth]{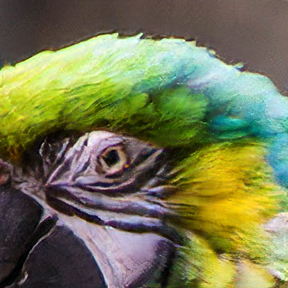}&
  	\includegraphics[width=.140\textwidth]{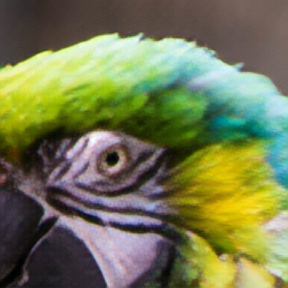}&
    \includegraphics[width=.140\textwidth]{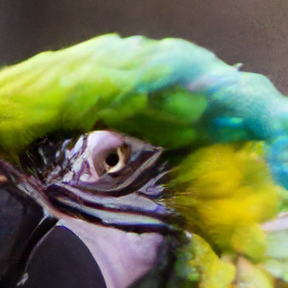}&
 	\includegraphics[width=.140\textwidth]{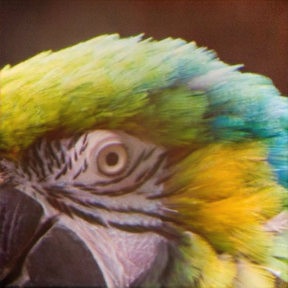}&
  	\includegraphics[width=.140\textwidth]{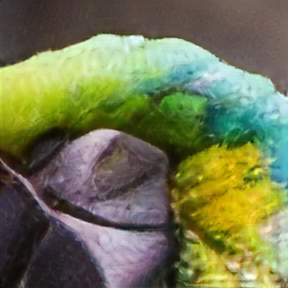}\\
  Ref.&Dis.1&Dis.2&Dis.3&Dis.4&Dis.5&Dis.6\\
   MOS$\uparrow$ &1359.45(1)&1327.90(2)&1261.15(3)&1213.73(4)&1206.27(5) &868.30(6)\\
PSNR$\uparrow$ &24.18(2) &22.99(4)&26.32(1)&23.61(3)&20.67(5) &19.91(6)\\
SSIM$\uparrow$ &0.679(3) &0.572(5)&0.720(2)&0.620(4)&0.863(1) &0.450(4)\\
MS-SSIM$\uparrow$ &0.893(3) &0.882(5)&0.934(2)&0.883(4)&0.938(1) &0.703(6)\\
LPIPS$\downarrow $ &0.198(4) &0.161(2)&0.174(3)&0.252(5)&0.110(1) &0.327(6)\\
IQT$\uparrow $ &1364.39(1) &1327.20(3)&1135.62(2)&1282.94(5)&1316.89(4)&1069.47(6)\\
Ours(SL)$\uparrow $ &0.765(1)&0.757(3)&0.758(2)& 0.734(5)&0.752(4)&0.689(6)\\
Ours(JSPL) $\uparrow$&0.765(1)&0.759(2)&0.756(3)&0.736(5)&0.754(4)&0.688(6)\\

\end{tabular}

\end{center}
\vspace*{-4mm}
\caption{An evaluation example from validation set of PIPAL. The quality is measured by MOS and 7 IQA methods. The numbers in brackets indicate the ranking of the corresponding distortion image.}
\label{fig:exam}
\vspace*{-6mm}
\end{figure*}

As shown in Fig.~\ref{fig:exam},  we take an example image from validation set of PIPAL to visually show the consistency between various methods and subjective perception, inlcuding PSNR, SSIM~\cite{SSIM}, MS-SSIM~\cite{MS-SSIM}, LPIPS~\cite{LPIPS}, IQT~\cite{IQT} and our method.
One can see that the proposed FR-IQA with JSPL achieves the closest rank agreement with the human annotated MOS.

\begin{table*}
  \centering
  \caption{Descriptions of the five IQA databases.}
\setlength{\tabcolsep}{1.6mm}{
\begin{tabular}{c|m{4.0em}<{\centering}|c|m{32.365em}<{\centering}}
\hline
Database & \multicolumn{1}{m{4.0em}<{\centering}|}{\# Ref.} & \multicolumn{1}{m{3.5em}<{\centering}|}{\# Dis.} & Distortion Types \bigstrut\\
\hline
TID2013~\cite{TID2013} & 25    & 3,000  & (1) Additive Gaussian noise; (2) Additive noise in color components;
(3) Spatially correlated noise; (4) Masked noise; (5) High frequency noise; (6) Impulse noise; (7) Quantization noise; (8) Gaussian blur; (9) Image denoising; (10) JPEG compression; (11) JPEG2000 compression; (12) JPEG transmission errors; (13) JPEG2000 transmission errors; (14)
Non eccentricity pattern noise; (15) Local block-wise distortions of different intensity; (16) Mean shift (intensity shift); (17) Contrast change; (18) Change of color saturation; (19) Multiplicative Gaussian noise; (20) Comfort noise; (21) Lossy compression of noisy images; (22) Image color quantization with dither; (23) Chromatic aberrations;
(24) Sparse sampling and reconstruction \bigstrut\\
\hline
LIVE~\cite{LIVE}  & 29    & 982   & (1) JPEG compression; (2) JPEG2000 compression; (3) Additive white Gaussian noise; (4) Gaussian blur; (5) Rayleigh fast-fading channel distortion   \bigstrut\\
\hline
CSIQ~\cite{CSIQ}  & 30    & 866   & (1) JPEG compression; (2) JP2K compression; (3) Gaussian blur; (4) Gaussian white noise; (5) Gaussian pink noise; (6) Contrast change \bigstrut\\
\hline
KADID-10k~\cite{KADID}  & 81    & 10,125 & (1) Gaussian blur; (2) Lens blur; (3) Motion blur; (4) Color diffusion; (5) Color shifting; (6) Color quantization; (7) Color over-saturation; (8) Color desaturation; (9) JPEG compression; (10) JP2K compression; (11) Additive white Gaussian noise; (12) White with color noise;  (13) Impulse noise; (14) Multiplicative white noise; (15) DnCNN denoising algorithm; (16) Brightness changes; (17) Darken; (18) Shifting the mean; (19) Jitter spatial distortions; (20) Non-eccentricity patch; (21) Pixelate; (22) Quantization; (23) Color blocking; (24) Sharpness; (25) Contrast \bigstrut\\

\hline
PIPAL~\cite{PIPAL}  & 250 & 25,850 & (1) Median filter denoising; (2) Linear motion blur; (3) JPEG and JPEG 2000; (4) Color quantization; (5) Gaussian noise; (6) Gaussian blur; (7) Bilateral filtering; (8) Spatial warping; (9) Comfort noise; (10) Interpolation; (11) A+; (12) YY;  (13) TSG; (14) YWHM; (15) SRCNN; (16) FSRCNN; (17) VDSR; (18) EDSR; (19) RCAN; (20) SFTMD; (21) EnhanceNet; (22) SRGAN; (23) SFTGAN; (24) ESRGAN; (25) BOE; (26) EPSR; (27) PESR; (28) EUSR; (29) MCML; (30) RankSRGAN; (31) DnCNN; (32) FFDNet; (33) TWSC; (34) BM3D; (35) ARCNN; (36) BM3D + EDSR; (37) DnCNN + EDSR; (38) ARCNN + EDSR; (39) noise + EDSR; (40) noise + ESRGAN;\bigstrut\\
\hline
\end{tabular}%

}
 \label{tab:list}%
\end{table*}%

\end{document}